\documentclass[sigconf]{acmart}
\usepackage{subcaption}
\usepackage{multirow}
\usepackage{colortbl}
\usepackage{threeparttable}
\usepackage{makecell}
\usepackage{tabularx}
\usepackage{longtable}

\usepackage{mathtools}      % 会自动加载amsmath，但不会冲突
\usepackage{pifont}

\usepackage[capitalize,noabbrev]{cleveref}

\theoremstyle{plain}

\theoremstyle{definition}

\theoremstyle{remark}

\AtBeginDocument{%
  }

\copyrightyear{2026}
\acmYear{2026}
\setcopyright{cc}
\setcctype{by}
\acmConference[KDD '26]{Proceedings of the 32nd ACM SIGKDD Conference on Knowledge Discovery and Data Mining V.2}{August 09--13, 2026}{Jeju Island, Republic of Korea}
\acmBooktitle{Proceedings of the 32nd ACM SIGKDD Conference on Knowledge Discovery and Data Mining V.2 (KDD '26), August 09--13, 2026, Jeju Island, Republic of Korea}
\acmDOI{10.1145/3770855.3818883}
\acmISBN{979-8-4007-2259-2/2026/08}
\begin{document}

%%
%% The "title" command has an optional parameter,
%% allowing the author to define a "short title" to be used in page headers.
\title{MedTVL: Harnessing Vision and Language for Medical Time Series Classification}

%%
%% The "author" command and its associated commands are used to define
%% the authors and their affiliations.
%% Of note is the shared affiliation of the first two authors, and the
%% "authornote" and "authornotemark" commands
%% used to denote shared contribution to the research.
\author{Jiexia Ye}
\email{jye324@connect.hkust-gz.edu.cn}
\affiliation{%
  \institution{The Hong Kong University of Science and Technology (Guangzhou)}
  \department{Data Science and Analytics Thrust} % 新增的
  \city{Guangzhou}
  \country{China}
}

\author{Jia Li}
\authornote{Corresponding Author} % 添加这个命令会在作者名字上方生成一个星号 *
\email{jialee@hkust-gz.edu.cn}
\affiliation{%
  \institution{The Hong Kong University of Science and Technology (Guangzhou)}
  \department{Data Science and Analytics Thrust} % 新增的
  \city{Guangzhou}
  \country{China}
}

\author{Fugee Tsung}
\email{season@ust.hk}
\affiliation{%
  \institution{The Hong Kong University of Science and Technology}
    \department{Department of Industrial Engineering \& Decision Analytics}
  \city{Hong Kong SAR}
  \country{China}
}

%%
%% By default, the full list of authors will be used in the page
%% headers. Often, this list is too long, and will overlap
%% other information printed in the page headers. This command allows
%% the author to define a more concise list
%% of authors' names for this purpose.
\renewcommand{\shortauthors}{Jiexia Ye, Jia Li \& Fugee Tsung}

\begin{abstract}
Recent advancements in multimodal learning for medical time series (MedTS) classification  highlight the benefits of integrating complementary modalities for clinical decision.
However, existing methods typically focus on bi-modal interactions (e.g., time series and text), leaving the tri-modal synergy between time series, vision, and language largely unexplored.
Inspired by diagnostic practice synergizing numerical assessment, visual inspection and clinical context, we introduce MedTVL, a text-guided dual-pathway architecture tailored for MedTS classification.
Specifically, it synergizes a convolution-based temporal pathway for fine-grained temporal dynamics from raw numerical sequences and a transformer-based visual pathway for holistic morphological structures from time-series-derived images. 
Such combination of cross-modal and architectural heterogeneity provides a comprehensive diagnostic perspective.
To further resolve potential diagnostic ambiguity, both pathways are guided by adaptive medical textual semantics.
Finally, a Mixture-of-Experts mechanism dynamically routes each instance to specialized fusion experts, capturing instance-specific reliance on the temporal and visual pathway outputs.
In addition, MedTVL supports multimodal contrastive learning to mitigate the clinical label scarcity challenge.
Extensive experiments across multiple medical datasets and tasks, spanning supervised, few-shot, and contrastive learning settings, demonstrate the superiority and transferability of MedTVL, highlighting its potential for robust clinical decision support.
\end{abstract}

%%
%% The code below is generated by the tool at http://dl.acm.org/ccs.cfm.
%% Please copy and paste the code instead of the example below.
%%


%%
%% This command processes the author and affiliation and title
%% information and builds the first part of the formatted document.
\maketitle

\section{Introduction}
%\begin{wrapfigure}{r}{0.45\textwidth}  % "r" 表示图像在右侧，0.5\textwidth 表示图像宽度为页面宽度的一半
\begin{figure}[htb]
  %\vspace{-3mm}
  \begin{center}
    \includegraphics[width=0.47\textwidth]{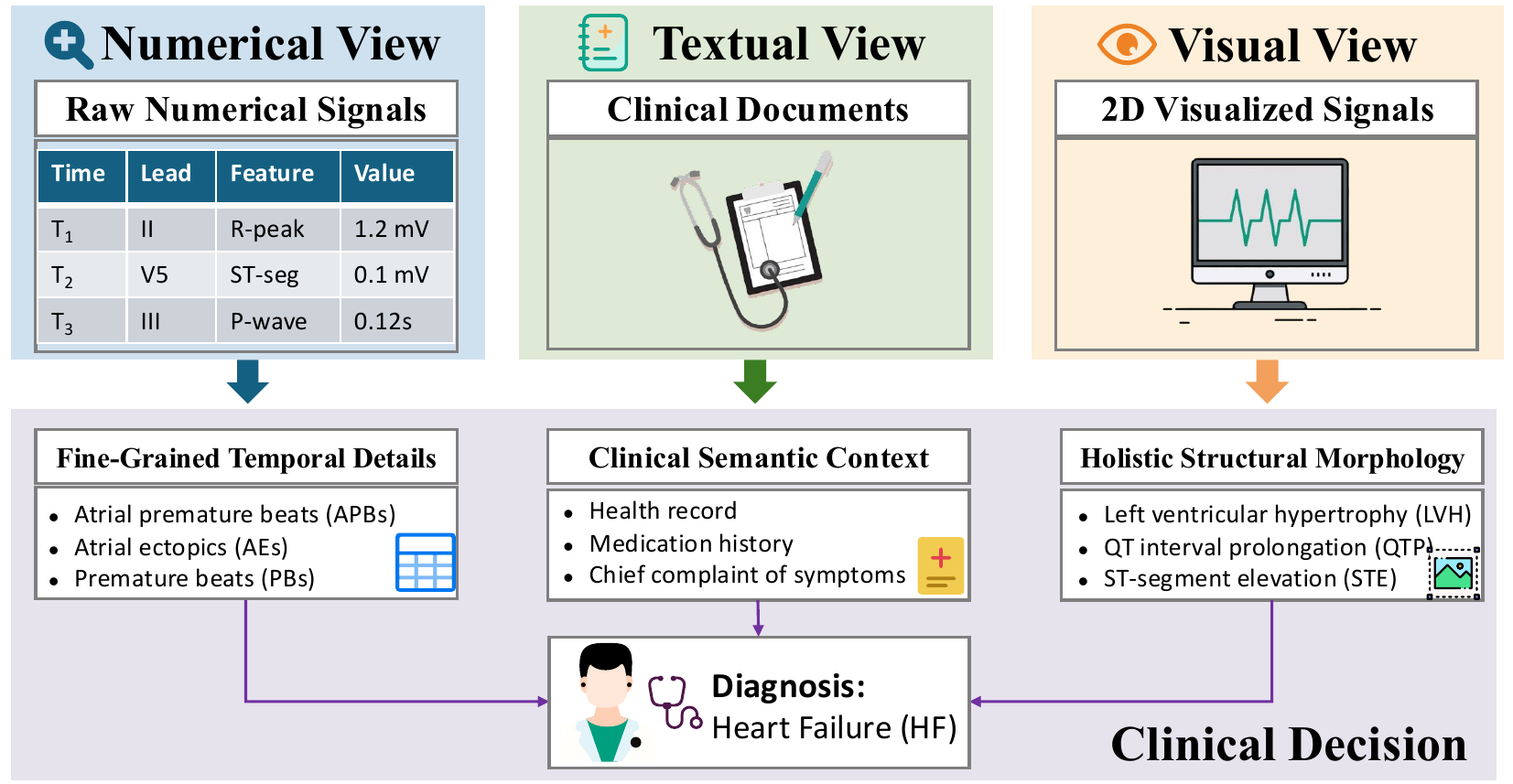} % 这里调整图像的宽度，确保它在wrapfigure中居中显示
  \end{center}
  \vspace{-3mm}
  \caption{Illustration of the multi-view diagnostic process for MedTS, highlighting the complementary strengths of numerical, visual, and textual views to support clinical decisions.}
    \label{fig:intro}
    \vspace{-3mm}
%\end{wrapfigure}
\end{figure}
Medical time series (MedTS) are widely applied in critical healthcare scenarios such as clinical monitoring, disease screening, and diagnostic decision-making \cite{gu2025foundation}, encompassing diverse physiological signals including electrocardiograms (ECG) \cite{ding2025advances}, electroencephalograms (EEG) \cite{sharma2024emerging}, and other vital signs \cite{fatourechi2007emg}.
The nature of MedTS is complex across several dimensions: First, significant data heterogeneity, characterized by multi-source signal modalities, disparate sampling rates, and diverse diagnostic tasks \cite{rim2020deep, woo2024unified}. Second, prominent multi-scale patterns, where certain patterns manifest as transient fluctuations within short time windows, while others are reflected in holistic morphological structures \cite{Medformer}. Third, inherent label scarcity, as incomplete annotations in real-world clinical environments often compromise the generalizability of models \cite{METS}. These multifaceted complexities pose challenges to developing robust and generalizable models for MedTS classification.

Traditional methods mostly rely on unimodal numerical signals \cite{Medformer, MedSpaformer}, with a few attempts to convert time series into images \cite{wu2024srt}.
However, unimodal modeling struggles to capture the full spectrum of diagnostic evidence, leading to sub-optimal performance.
With the advancement of multimodal learning, an increasing number of studies have integrated time series with text, validating the significance of textual semantics in supporting clinical decision-making \cite{MERL, MedTsLLM}.
However, due to the lack of explicit visual modeling, these methods fail to capture critical morphological patterns, which deviates from clinical practice where clinicians heavily rely on visual reasoning.
Recently, time-series-vision fusion has emerged within the general time series domain \cite{DMMV, OccamVTS}; however, these methods lack textual grounding and are not tailored to the unique properties of medical data. While interest in tri-modal modeling (time series, vision, and language) is growing, existing studies remain scarce and are restricted to either general-purpose forecasting \cite{Time-VLM} or specific medical modalities \cite{GEM}. For instance,  TimeVLM \cite{Time-VLM} focuses on time series forecasting with temporal-centric attention, treating vision and text as auxiliary, a limitation that prevents full exploitation of these modalities. GEM \cite{GEM}, specifically designed for ECG, maps time series and images into a shared textual space, which may buffer and dilute fine-grained physiological details. 
%Both adopt coarse-grained modality-level fusion, overlooking sample-level heterogeneity. Thus, they may lack the flexibility to handle the diversity of medical signals.

In real-world clinical practice, the diagnosis of MedTS is inherently a multi-view process \cite{GEM, MedViA}. Clinicians often examine raw temporal signals to identify fine-grained pathological patterns (e.g. premature ventricular contractions \cite{klewer2022premature}). Complementarily, visual inspection of signal enables them to assess holistic morphology (e.g.,ST-segment elevation \cite{mclaren2024st}).  Meanwhile, clinical text provides essential semantic context by incorporating patient history and prior assessments, allowing clinicians to confirm, refine, or revise their judgments. Together, these views reflect how clinicians integrate precise temporal details, global visual cues, and semantic information into a coherent diagnostic decision.

Building upon and transcending these clinical practices, we propose \textbf{MedTVL}, a unified framework that integrates \textbf{T}ime series, \textbf{V}ision, and \textbf{L}anguage, tailored for MedTS classification. 
MedTVL employs a text-guided dual-pathway architecture to model diverse diagnostic patterns across multiple scales.
Specifically, the temporal pathway utilizes a convolutional backbone to extract fine-grained dynamics from raw numerical signals, while the visual pathway adopts a transformer-based backbone to capture holistic morphology from time series-derived Continuous Wavelet Transform (CWT) images.
This synergistic design goes beyond modality-level complementarity by explicitly introducing architectural heterogeneity, where convolutional and transformer-based backbones offer complementary inductive biases for multi-scale diagnostic modeling.
Subsequently, shared textual semantics are adaptively injected into the pathway subspaces to constrain modality-specific learning with clinical context.
Finally, to address the varying importance of temporal and visual cues across samples, MedTVL introduces a Mixture-of-Experts (MoE) mechanism that dynamically assigns each instance to specialized fusion experts. This fine-grained fusion allows the model to adaptively reconcile heterogeneous signals, providing robust and flexible integration across diverse clinical scenarios.
%To integrate these features, a Mixture-of-Experts (MoE) mechanism is introduced for sample-wise adaptive fusion. By dynamically weighting each pathway's contribution, MedTVL effectively handles medical data heterogeneity, enabling robust adaptation across diverse signal types and clinical tasks. 
Additionally, the dual-pathway structure naturally forms cross-modal positive pairs, facilitating multimodal contrastive learning to mitigate the label scarcity challenge in clinical settings.
% While the visual pathway captures the holistic spectral structure, the Inception-based temporal pathway ensures that point-wise numerical precision is preserved, forming a comprehensive dual-perspective analysis.
%By combining a CNN-based temporal pathway (focusing on shift-invariant local motifs) with a Transformer-based visual pathway (modeling global spectral-spatial rhythms), our framework achieves a heterogeneous feature fusion. This design ensures the capture of both micro-temporal oscillations and macro-frequency dependencies, providing a comprehensive diagnostic perspective that no single-architecture approach could offer."
%Furthermore, this choice introduces architectural heterogeneity to the dual-pathway framework. While the temporal pathway leverages CNNs to focus on localized morphological shifts in raw tracings, the visual pathway utilizes Transformers to model holistic frequency-spatial rhythms. This synergy ensures that the model captures both the microscopic details of 1D waveforms and the macroscopic patterns of the 2D time-frequency plane, leading to a more robust and comprehensive diagnostic representation.

\begin{itemize}
\item
To the best of our knowledge, MedTVL is the first tri-modal framework that encapsulates clinical diagnostic perspectives for general-purpose medical time series classification.
\item
MedTVL introduces a dual-pathway architecture to capture multi-scale patterns from synergistic temporal-visual perspectives, using adaptive textual guidance and MoE-based instance-level fusion to handle medical data heterogeneity. It also supports multimodal contrastive learning.

%MedTVL introduces a dual-pathway architecture to mimic clinical practices by synergizing global morphological trends via a text-augmented visual pathway with local temporal dynamics via a textual-guided temporal pathway. An MoE-based mechanism further enables sample-wise adaptive heterogeneous fusion to handle clinical data variability.

\item
We conduct comprehensive experiments on datasets spanning multiple medical tasks, covering supervised, few-shot, and contrastive learning settings. Our results demonstrate that MedTVL outperforms SOTA baselines, while also showcasing its capability to address the label scarcity challenge. The code link is \text{https://github.com/start2020/MedTVL}

\end{itemize}
% \begin{figure*}[htb]
\begin{figure*}[ht] % 强制图片放在当前页面顶部
  \centering
  \includegraphics[width=1.0\textwidth]{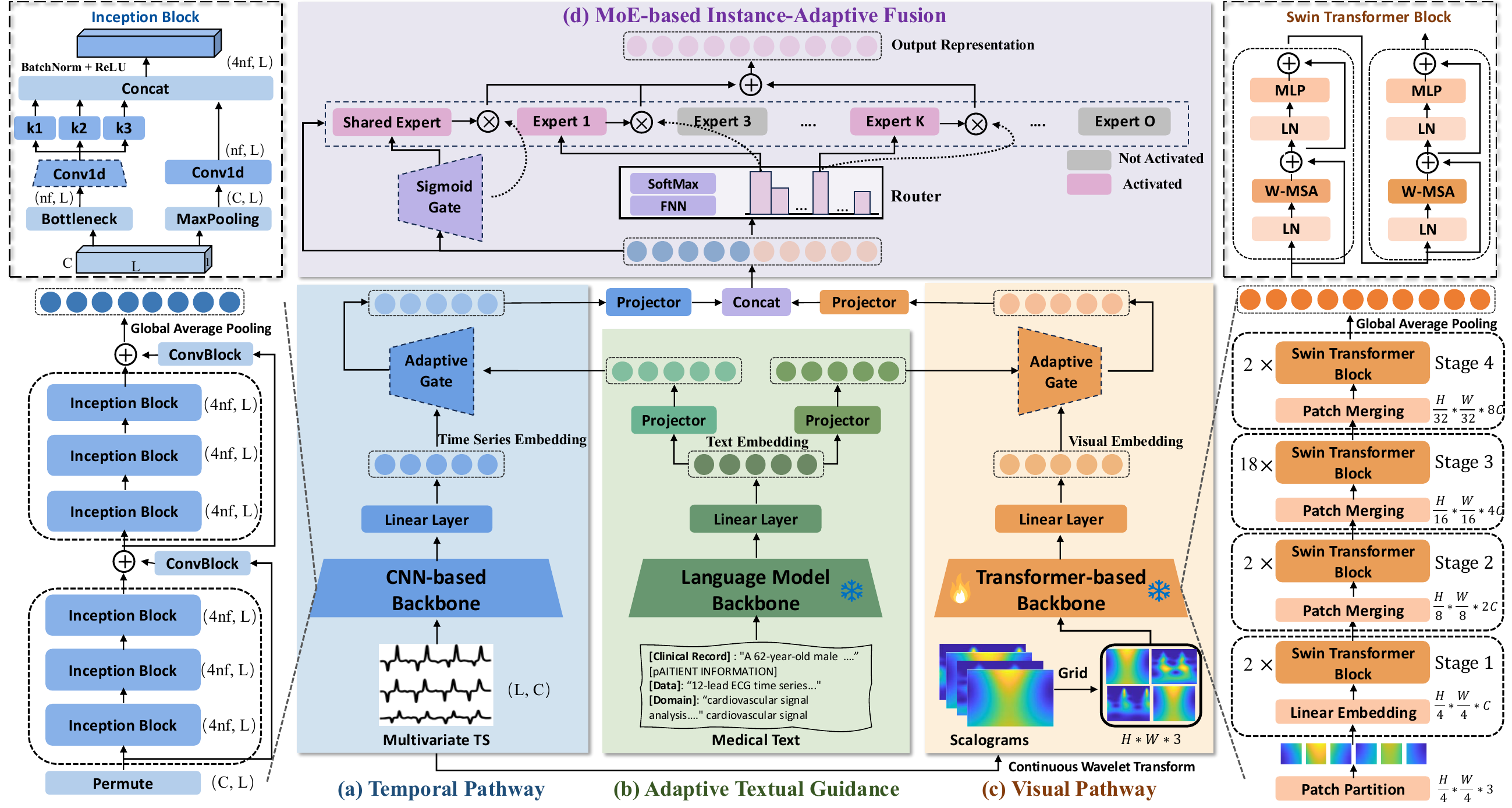}
  \vspace{-4mm}
  \caption{
Overview of the MedTVL framework. More details are in section \ref{Methodology}.
  }
  \label{fig:model}
\vspace{-3mm}
\end{figure*}

%(b) The time series encoder utilizes multi-granularity encoding to capture multi-scale temporal patterns and cross-channel encoding to capture channel interactions. 
%  (c)  We implement multi-granularity encoding with multiple TSDA blocks, which capture both intra-granularity and inter-granularity correlations.  

\section{Related Work}
%Here, we focus on time series representation learning methods that benefit MedTS classification.
%Here, we focus on multimodal time series representation learning methods that benefit MedTS classification. A detailed discussion of unimodal approaches and contrastive learning for time series is provided in the \textbf{Appendix}.
\textbf{Unimodal Medical Time Series Classification.} 
Unimodal approaches predominantly rely on numerical medical signal modeling, with a limited number of studies exploring image-based representations of time series \cite{wu2024srt, pratiher2022dilated}. Many recent deep learning approaches adopt architectures such as convolutional neural networks (CNNs) \cite{Lawhern_2018}, recurrent neural networks (RNNs) \cite{Salloum_Kuo_2017}, graph neural networks (GNNs) \cite{Tang2021SelfSupervisedGN}, and Transformers \cite{Medformer, MedSpaformer}. While such methods have shown promising results, they are often tailored for specific signals (e.g., ECG) with limited cross-signal generalization. Moreover, relying on a unimodal perspective constrains their capacity to fully capture the variability and heterogeneity of real-world clinical data, thereby motivating increasing interest in multimodal learning.

\textbf{Bi-modal Time Series Representation Learning}
Current multimodal research focuses predominantly on time-series–text alignment, particularly within general time-series domains \cite{cheng2024advancing, ye2024empowering} (e.g. Time-LLM \cite{jin2023timellm}, UniTS \cite{gao2024units}). 
In the medical domain, some works leverage Large Language Models (LLMs) to integrate time series and text (e.g., MedTsLLM \cite{MedTsLLM}, MedualTime \cite{MedualTime}) for supervised learning while other focus on report-guided ECG contrastive learning \cite{METS,MERL}.
However, these methods typically overlook explicit visual modeling, failing to capture clinically meaningful morphological patterns essential for diagnostic practice.
Although time series–vision approaches have recently emerged \cite{DMMV, OccamVTS}, they primarily focus on forecasting and remain unoptimized for the unique challenges of clinical diagnostics.
A notable exception is MedViA \cite{MedViA}, which is specifically designed for medical data; however, it relies on static fusion and lacks clinical semantic guidance, limiting its flexibility to handle MedTS variability. 

\textbf{Tri-modal Time Series Representation Learning.} 
The exploration of tri-modal frameworks remains more scarce, to our knowledge, with only Time-VLM \cite{Time-VLM} and GEM \cite{GEM} reported to date.
Time-VLM \cite{Time-VLM} prioritizes forecasting via temporal-dominant attention, limiting full exploitation of the visual and textual inputs. Its gated fusion applies a shared weighting across instances, potentially limiting the model’s flexibility in integrating heterogeneous pathway outputs.
Conversely, GEM \cite{GEM} collapses both numerical signals and visual images into a unified textual space, which inevitably erodes fine-grained waveform dynamics and morphological details. %Furthermore, both methods employ a one-fits-all fusion strategy, lacking instance-level adaptability.
%MedTVL addresses these gaps through visual-temporal primary modeling, textual adaptive augmentation, and adaptive sample-level modality fusion.

\textbf{Contrastive Learning for Time Series.} 
% (e.g., CoST \cite{woocost}, TFC \cite{zhang2022self})
%Contrastive learning is crucial for addressing clinical label scarcity, yet conventional approaches typically focus on data augmentations, failing to capture meaningful clinical morphological features. 
Contrastive learning is a powerful paradigm for mitigating label scarcity. In the general time series domain, it has evolved from methods defining positive pairs \cite{ tonekaboniunsupervised}, to augmentation-based multi-view alignment (e.g., TS-TCC \cite{TS-TCC}, TS2Vec \cite{Ts2vec}), and further to approaches exploiting intrinsic temporal properties, such as frequency-domain consistency and seasonal–trend disentanglement \cite{woocost,zhang2022self}. Medical time series contrastive learning has progressed from early adoption of general-domain techniques to deeper adaptation to medical characteristics while most of them are tailored to specific signal types (e.g., EEG) \cite{liu2023self}. All these methods above remain confined to the time series modality. Recently, some works focus on report-guided ECG contrastive learning \cite{METS,MERL}. However, the absence of paired text in most public  datasets limits the application of report-guided contrastive learning. AimTS \cite{AimTS}, a recent pioneering work, introduces time series–image contrastive learning to enhance representation generalization. Compared with MedTVL, it is designed for general time series and lacks textual semantic support.

\section{Methodology}
\label{Methodology}
\textbf{Problem Formulation.}
Consider a medical dataset with $N$ samples $\mathcal{D} = \{(\mathbf{X}_i, s_i, y_i)\}_{i=1}^N$, where $\mathbf{X} \in \mathbb{R}^{L \times C}$ denotes the raw temporal signal with length $L$ and $C$ channels; $y \in \mathcal{Y} = \{1, 2, \dots, M\}$ is its label and $M$ is the number of classes; $s_i$ represents its associated clinical semantics. 
To ensure broad applicability across clinical settings, $s$ is defined with flexible granularity, ranging from sample-level diagnostic reports to dataset-level semantic descriptions, or their combination. For $\mathbf{X}$, a corresponding visual representation $\mathbf{V} = g(\mathbf{X})$ is derived. Our objective is to develop a  framework $f(\cdot)$ that leverages the complementary strengths of the tri-modal information for MedTS classification: $\bar{y} = f(\mathbf{X}, \mathbf{V}, s; \Theta)$ where $\Theta$ denotes the trainable parameters. The framework under supervised learning is optimized to by minimizing the cross-entropy loss: $\mathcal{L}_{ce} = -\frac{1}{N} \sum_{i=1}^{N} y_i \log(\hat{y}_i)$.

\textbf{Overview.}
MedTVL is a tri-modal framework inspired by the holistic diagnostics in clinical practice, designed for comprehensive medical time-series (MedTS) modeling. 
As illustrated in Figure 1, MedTVL comprises four key components.
The temporal pathway adopts InceptionTime \cite{InceptionTime} as backbone to process raw time series and extract multi-scale patterns from a numerical perspective. In parallel, the visual pathway employs a Swin Transformer to capture structural representations from CWT scalograms, complementing temporal dynamics with morphological cues. 
%It employs the Continuous Wavelet Transform (CWT) to transform 1D signals into time-frequency spectrograms, which are then processed by a Swin Transformer to model global spectral-spatial rhythms.
Within each pathway, shared textual information is adaptively projected into modality-specific representations to provide clinical semantic guidance. 
Finally, the text-enhanced dual-pathway features are routed through a Mixture-of-Experts module, enabling adaptive fusion of temporal and visual cues on a per-sample basis.
%Details of contrastive learning with MedTVL are presented in Section~\ref{Contrastive}.

\subsection{Convolutional Network-based Temporal Pathway}
The temporal pathway aims to characterize the fine-grained multi-scale temporal dynamics from a numerical perspective. 
%It utilizes InceptionTime \cite{} to process 1D signals across multi-scale receptive fields, enabling the extraction of fine-grained temporal features and local morphological transients.
Medical time series often exhibit clinically informative patterns that are localized in time and vary in temporal extent, such as brief waveform distortions or short-lived abnormal rhythms \cite{liu2023self}. Convolutional neural networks \cite{cui2016multi, zhao2017convolutional} are well suited for modeling such intrinsic numerical dynamics due to their strong locality bias and temporal shift invariance.

\textbf{InceptionTime as Backbone.}  Among CNN-based architectures, we adopt InceptionTime \cite{InceptionTime} as the backbone of the temporal pathway, owing to its ability to model temporal patterns across multiple receptive fields simultaneously. Unlike conventional 1D CNNs with fixed kernel sizes \cite{he2016deep, wang2017time}, InceptionTime employs parallel convolutional filters with different temporal spans, allowing it to robustly capture pathological patterns occurring at diverse time scales, which is particularly important for heterogeneous medical signals. Moreover, InceptionTime incorporates bottleneck convolutions to reduce computational complexity and residual connections to facilitate stable optimization, making it well suited for long medical time series and data-limited clinical settings. For brevity, we present a simplified formulation of the InceptionTime encoder, focusing on its multi-scale integration:
%\vspace{-2mm}
\begin{equation}
\mathbf{h} = \text{GAP} \left( \sigma \left( \text{Concat}_{k \in \mathcal{K} \cup \{p\}} (\mathbf{Z}_k) + \phi(\mathbf{X}) \right) \right)
\end{equation}
where $\text{GAP}(\cdot)$ denotes the global average pooling; $\sigma(\cdot)$ refers to ReLU; $\mathcal{K}$ represents the set of parallel convolutional kernel sizes; $\mathbf{Z}_k= \text{Conv}_{k} (\text{Conv}_{1}(\mathbf{X}) ) $ is the feature map generated by the $k$-th scale branch, while $\mathbf{Z}_p= \text{Conv}_{1} ( \text{MaxPool}(\mathbf{X})) $ signifies the max-pooling branch; $\phi(\mathbf{X})$ denotes the residual connection. $\mathbf{h}_{ts}\in \mathbb{R}^{D}=f_{\text{MLP}}(\mathbf{h})$ denotes the final embedding of temporal pathway.

Although InceptionTime can aggregate local features into a global representation, such aggregation is intrinsically driven by localized temporal patterns, leaving global structural morphology under-modeled at the temporal pathway, thereby motivating a complementary visual pathway.

% While the visual pathway captures the holistic spectral structure, the Inception-based temporal pathway ensures that point-wise numerical precision is preserved, forming a comprehensive dual-perspective analysis.
%By combining a CNN-based temporal pathway (focusing on shift-invariant local motifs) with a Transformer-based visual pathway (modeling global spectral-spatial rhythms), our framework achieves a heterogeneous feature fusion. This design ensures the capture of both micro-temporal oscillations and macro-frequency dependencies, providing a comprehensive diagnostic perspective that no single-architecture approach could offer."
%Furthermore, this choice introduces architectural heterogeneity to the dual-pathway framework. While the temporal pathway leverages CNNs to focus on localized morphological shifts in raw tracings, the visual pathway utilizes Transformers to model holistic frequency-spatial rhythms. This synergy ensures that the model captures both the microscopic details of 1D waveforms and the macroscopic patterns of the 2D time-frequency plane, leading to a more robust and comprehensive diagnostic representation.

\subsection{Transformer-based Visual Pathway}
The visual pathway is designed to mirror and deepen clinical visual inspection to model global spectral-spatial rhythms.

\textbf{CWT-based Visual Representation.} 
Converting time series into images has gained increasing attention in recent time-series studies, including line plots \cite{ViTST}, heatmaps \cite{VisionTS}, and spectrograms derived from the Short-Time Fourier Transform (STFT) \cite{dixit2024vision} and the Continuous Wavelet Transform (CWT) \cite{almanza2023emotion}. In practice, clinicians primarily rely on 1D tracings (i.e., line plots) to identify physiological patterns. However, subtle pathological signals are often obscured by complex temporal fluctuations. While STFT-based spectrograms are constrained by their fixed time–frequency resolution, limiting its ability to capture global spatiotemporal correlations, the Continuous Wavelet Transform (CWT) inherently provides a multi-resolution analysis. This property allows CWT to preserve global structural information while effectively capturing transient spectral variations. 
In this paper, we apply CWT to process the raw time series data. Specifically, 
for a multivariate time series sample $\mathbf{X}$ with $C$ channels,
we first apply CWT to each channel independently, producing a set of
scalograms $\mathcal{V}_i = CWT(\mathbf{X})=\{ v_{i,1}, v_{i,2}, \dots, v_{i,C} \}$,
where each $v_{i,c} \in \mathbb{R}^{h \times w}$ denotes the time-frequency representation of the $c$-th channel. 

\textbf{Grid-based Image Creation.}
Furthermore, inspired by \cite{ViTST}, we arrange the scalograms of all channels into a structured grid following the standard clinical ordering of leads. This layout facilitates the modeling of inter-channel correlations, closely mirroring the way clinicians cross-reference diagnostic patterns across multiple leads. 
Specifically, each scalogram is first rendered as a three-channel image by applying a fixed colormap.
The rendered scalogram images are then arranged into a single grid-structured image $\mathbf{V}_i = \mathcal{G}(\mathcal{V}_i)$ using a
predefined grid layout.
By default, we adopt a square grid and organize the $C$ channel-wise
scalograms into a grid of size $l \times l$ when $(l-1)^2 < C \leq l^2$,
with unused grid positions left empty.
To ensure structural consistency across samples, the grid layout and channel ordering are fixed within each dataset. This grid-based construction preserves channel-wise independence while enabling cross-channel interactions to be implicitly modeled by visual backbones. 

%Figure \ref{fig:grid} shows an example of gird based line plot, CWT spectrogram and STFT spectrogram on two datasets.

To summarize, the CWT-based grid visual representation $\mathbf{V}$ is constructed as follows:
%\vspace{-3mm}
\begin{equation}
\mathbf{V} = \mathcal{G}(\text{CWT}(\mathbf{X}))
\end{equation}
where  $\mathcal{G}(\cdot)$ is grid mapping operation. $\mathbf{X}$ is the raw multi-channel time series and $\mathbf{V} \in \mathbb{R}^{H \times W \times 3}$ is its 2D grid image.

\textbf{Swin Transformer as Backbone.} 
To extract latent embeddings from the 2D scalograms, we employ the Swin Transformer \cite{liu2021swin} as vision backbone. 
Unlike conventional CNNs \cite{he2016deep, huang2016deep} that primarily capture local patterns or standard Vision Transformers \cite{han2021transformer, ViT} that suffer from quadratic complexity, the Swin Transformer provides a hierarchical architecture with shifted window partitioning, aligning with the intrinsic properties of CWT scalograms.
First, its progressive downsampling produces hierarchical feature maps, supporting multi-scale modeling of both transient and macro-level patterns. 
Second, its local window multi-head self-attention (W-MSA) ensures efficient computation, while its shifted version (SW-MSA) enables cross-window interactions to capture long-range spatial-frequency correlations across the CWT grid. 
Together, W-MSA and SW-MSA form consecutive Swin Transformer blocks:
%\vspace{-2mm}
\begin{equation}
\begin{aligned}
&\bar{\mathbf{h}}^l = \text{W-MSA}(\text{LN}(\mathbf{h}^{l-1})) + \mathbf{h}^{l-1}, \\
&\mathbf{h}^l = f_\text{MLP}(\text{LN}(\bar{\mathbf{h}}^l)) + \bar{\mathbf{h}}^l, \\
&\bar{\mathbf{h}}^{l+1} = \text{SW-MSA}(\text{LN}(\mathbf{h}^l)) + \mathbf{h}^l, \\
&\mathbf{h}^{l+1} = f_\text{MLP}(\text{LN}(\bar{\mathbf{h}}^{l+1})) + \bar{\mathbf{h}}^{l+1}
\end{aligned}
\label{eq:swin_blocks}
\end{equation}
where $l$ is the block index. $\text{LN}$ and $f_\text{MLP}$ denote normalization and fully-connected layers. Finally, the pooled visual features derived from Swin Transformer are projected into a lower-dimensional latent space and normalized as $\mathbf{h}_{\text{img}} \in \mathbb{R}^{D}$.
Notably, MedTVL is flexible, supporting alternative transformer-based vision backbones.

\subsection{Adaptive Textual Guidance}
\label{subsec:text}
In clinical practice, medical text is often used as auxiliary context to disambiguate numerical signals and imaging findings \cite{MedTsLLM}. 
Motivated by this, we leverage medical text to guide dual-pathway learning, aligning representations with clinical semantics and mitigating overfitting to superficial patterns. 
However, due to the heterogeneous distributions of temporal and visual modalities, directly sharing a single textual embedding may cause feature degradation. 
Thus, we propose adaptive textual guidance that projects shared text into modality-specific subspaces and integrates it into each pathway via gated fusion, enabling modality-aware constraints.

Specifically, a frozen medical language model encodes the text $s$ into a shared embedding,
$\mathbf{h}_{\text{txt}} = f_{\text{LM}}(s)$.
This embedding is projected into pathway-specific subspaces,
$\mathbf{h}^{p}_{\text{txt}} \in \mathbb{R}^{D} = f_{\text{MLP}}^{p}(\text{LN}(\mathbf{h}_{\text{txt}}))$, $p \in \{\text{ts}, \text{img}\}$. To selectively integrate textual evidence, 
a adaptive gating vector is computed as
$\mathbf{g} \in \mathbb{R}^{D} = \sigma(\mathbf{W}_g [\mathbf{h}_{p}; \mathbf{h}^{p}_{\text{txt}}])$,
and the enhanced representation $\bar{\mathbf{h}}_{p}$ is obtained via  weighted fusion:
%\vspace{-2mm}
\begin{equation}
\bar{\mathbf{h}}_{p} = (1 - \mathbf{g}) \odot \mathbf{h}_p + \mathbf{g} \odot \mathbf{h}^{p}_{\text{txt}}
\end{equation}
where $\odot$ denotes element-wise multiplication. For temporal pathway $\bar{\mathbf{h}}_p = \bar{\mathbf{h}}_{\text{ts}}$ and for visual pathway $\bar{\mathbf{h}}_p = \bar{\mathbf{h}}_{\text{img}}$. 
By injecting pathway-specific clinical semantics, the module provides enriched representations for the subsequent MoE fusion, facilitating instance-level expert selection.

\subsection{MoE-based Instance-Adaptive  Fusion}
%Traditional methods like simple concatenation or gate fusion, often fail to capture such sample-wise variability, leading to suboptimal fusion. Mixture of Experts (MoE) \cite{mu2025comprehensive} has recently been adopted in time-series modeling to enable adaptive specialization through expert selection \cite{liumoirai,liu2025mofe}. Such mechanism is particularly suitable for sample-level heterogeneous fusion to handle MedTS variability. 

The inherent diversity of medical samples poses a significant challenge for dual-pathway fusion, as diagnostic cues often prioritize either temporal dynamics or morphological patterns, depending on the underlying pathology \cite{yun2025temporal}. 
While simple concatenation is static and gated fusion relies on shared weights, these methods may struggle to capture complex interactions across heterogeneous temporal and visual features, potentially leading to suboptimal integration in MedTS.
%While simple concatenation are inherently static and gated fusion typically modulates modality contributions through a shared gating mechanism, these approaches may have limited capacity to capture the intricate, high-dimensional sample-wise variability present in MedTS, potentially leading to suboptimal fusion.
Mixture of Experts (MoE) \cite{mu2025comprehensive} has recently been introduced into time-series modeling to enable fine-grained adaptive specialization via parameter-level expert selection \cite{liumoirai,liu2025mofe}. By routing individual instances to specialized sub-networks, MoE provides a more expressive mechanism for heterogeneous fusion, allowing diverse fusion patterns to be modeled by distinct expert parameters. Following \cite{Time-MoE}, we reconfigure a shared pool of sparsely activated experts for instance-adaptive pathway-level fusion.

Specifically, we first apply modality-specific projection to align heterogeneous pathways into a unified space: $\mathbf{z}_{\text{img}} = f_{\text{MLP}}^{\text{img}}(\text{LN}(\bar{\mathbf{h}}_{\text{img}}))$ and $\mathbf{z}_{\text{ts}} = f_{\text{MLP}}^{\text{ts}}(\text{LN}(\bar{\mathbf{h}}_{\text{ts}}))$. 
The projected representations are then concatenated as a joint embedding $\mathbf{z} = [\mathbf{z}_{\text{img}}; \mathbf{z}_{\text{ts}}]$, which serves as the input to MoE fusion module. This module consists of a shared expert $\text{E}_{\text{shared}}(\cdot)$ for modeling global physiological patterns shared across instances and $O$ experts ${\text{E}_i(\cdot)}_{i=1}^{O}$ to for capturing instance-specific variations.
In practice, each expert mirrors the architecture of a standard FFN \cite{mu2025comprehensive}. A sigmoid gate $\sigma(W_s \mathbf{z})$ controls the contribution of the shared expert, and a sparse gating function $\text{G}(\mathbf{z})$ selects the top-$K$ routed experts.
The fused representation $\bar{\mathbf{z}}$ is denoted as:
\vspace{-2mm}
\begin{equation}
\bar{\mathbf{z}} = \sigma(W_s \mathbf{z}) \cdot \text{E}_{shared}(\mathbf{z}) + \sum_{k=1}^{K} G(\mathbf{z})_k \cdot \text{E}_k(\mathbf{z})
\end{equation}
where $G(\mathbf{z}) = \text{TopK}(\text{Softmax}(W_g \mathbf{z}))$ represents the gating weights that route inputs to the top $K$ most relevant specialized sub-networks. It effectively mitigates pathway conflicts and enables MedTVL to prioritize the most informative features based on the sample properties. Finally, the fused representation $\bar{\mathbf{z}}$ is fed into a MLP classifier to produce the final diagnostic predictions: $\hat{y} = f_\text{MLP} (\bar{\mathbf{z}})$. 

%where $\text{E}_{shared}$ captures universal physiological patterns via a sigmoid-gated mechanism, and $\{\text{E}_k\}$ are $N$ sparse experts that specialize in distinct clinical sub-populations. By treating each sample as an independent token for MoE routing, MedTVL achieves automatic instance-level specialization, ensuring robust fusion across diverse and complex medical scenarios.

%where $\text{E}_{shared}$ is a designated shared expert that consolidates universal physiological knowledge across all instances. Concurrently, a trainable gating network $G(\mathbf{z}_i)$ routes the input to the top-$k$ most relevant independent experts $\{\text{E}_k\}$, allowing the model to dynamically prioritize temporal or visual nuances based on individual patient patterns. This dual-structured MoE effectively mitigates modality conflicts and enhances the representation's adaptivity to complex clinical scenarios.

\subsection{Multimodal Contrastive Learning}
\label{Contrastive}
MedTVL’s dual pathways provide aligned multimodal views of the same physiological signals, namely temporal dynamics and visual morphology. In this section, we describe how MedTVL is leveraged for multimodal contrastive learning.

\textbf{Dual-level Contrastive Loss.} To enhance robustness, we employ a dual-level contrastive objective: an inter-modality loss to harmonize cross-modal semantics and an intra-modality loss to preserve individual discriminability. As detailed in Section \ref{subsec:text}, $\bar{\mathbf{h}}_{\text{ts}}^i$ and $\bar{\mathbf{h}}_{\text{img}}^i$ denote the text-enhanced embeddings of the temporal and visual pathways for sample $i$ within a batch. For the inter-modality objective, we treat the cross-pathway pair $(\bar{\mathbf{h}}_{\text{ts}}^i, \bar{\mathbf{h}}_{\text{img}}^i)$ as a natural positive pair, while instances $\bar{\mathbf{h}}_{\text{img}}^j$ (where $j \neq i$) serve as cross-modal negative samples. 
For the intra-modality objective, we construct positive pairs $(\bar{\mathbf{h}}_p^i, \bar{\mathbf{h}}_{p,\text{aug}}^i)$ using augmented views of the same modality and negative pairs ($\bar{\mathbf{h}}_p^i,\bar{\mathbf{h}}_p^j$) where $p \in \{ts, \text{img}\}$.
Weak stochastic augmentations are applied to generate positive pairs: jittering and scaling for the temporal pathway, and random cropping and flipping for the visual pathway.
The total contrastive loss $\mathcal{L}_{cl}$ is defined as:

\vspace{-2mm}
\begin{equation}
\mathcal{L}_{cl}
=
\mathcal{L}_{\text{inter}}\!\left(
\bar{\mathbf{h}}_{\text{ts}}, \bar{\mathbf{h}}_{\text{img}}
\right)
+
\lambda
\sum_{p \in \{\text{ts}, \text{img}\}}
\mathcal{L}_{\text{intra}}\!\left(
\bar{\mathbf{h}}_{p}, \bar{\mathbf{h}}_{p,\text{aug}}
\right)
\end{equation}
% \begin{equation}
% \mathcal{L}{cl} = \mathcal{L}_{\text{inter}}(\bar{\mathbf{h}}_{\text{ts}}, \bar{\mathbf{h}}_{\text{img}}) + \lambda \sum_{p \in \{\text{ts}, \text{img}\}} \mathcal{L}_{\text{intra}}(\bar{\mathbf{h}}_m, \bar{\mathbf{h}}_{m,\text{aug}})
% \end{equation}
where we employ the CLIP contrastive loss for both inter- and intra-modality objectives, and $\lambda$ is a trade-off coefficient balancing the two objectives and set as 0.5 in our paper. The CLIP contrastive loss is defined as follows:
\begin{equation}
\mathrm{sim}(\mathbf{h}^t_i, \mathbf{h}^v_j) = \frac{\mathbf{h}^t_i \cdot \mathbf{h}^v_j}{\tau}
\end{equation}

\begin{equation}
\mathcal{L}_{t \rightarrow v} = - \frac{1}{B} \sum_{i=1}^{B} 
\log \frac{\exp(\mathrm{sim}(\mathbf{h}^t_i, \mathbf{h}^v_i))}
{\sum_{j=1}^{B} \exp(\mathrm{sim}(\mathbf{h}^t_i, \mathbf{h}^v_j))}
\end{equation}

\begin{equation}
\mathcal{L}_{v \rightarrow t} = - \frac{1}{B} \sum_{i=1}^{B} 
\log \frac{\exp(\mathrm{sim}(\mathbf{h}^v_i, \mathbf{h}^t_i))}
{\sum_{j=1}^{B} \exp(\mathrm{sim}(\mathbf{h}^v_i, \mathbf{h}^t_j))}
\end{equation}

\begin{equation}
\mathcal{L}_{\mathrm{CLIP}} = \frac{1}{2} \Big( \mathcal{L}_{t \rightarrow v} + \mathcal{L}_{v \rightarrow t} \Big)
\end{equation}

where $B$ denotes the mini-batch size, $\mathbf{h}^t$ and $\mathbf{h}^v$ are $L_2$-normalized embeddings of the sample pair and $\tau$ is a temperature hyperparameter that scales the logits. The function $\text{sim}(\cdot, \cdot)$ denotes the cosine similarity. In this framework, $\mathcal{L}_{\text{inter}}$ encourages the alignment between the temporal and visual modalities by pulling $\bar{\mathbf{h}}_{\text{ts}}$ and $\bar{\mathbf{h}}_{\text{img}}$ closer in the joint embedding space. Conversely, $\mathcal{L}_{\text{intra}}$ facilitates modality-specific robustness by ensuring that the representations of raw data and its augmented versions remain consistent.

\textbf{Pre-training and Fine-tuning.} At the pre-training stage, we perform contrastive learning on the text-enhanced outputs of the dual pathways while discarding the MoE module; this design prevents the model from taking shortcuts (e.g., relying solely on features from one pathway) to complete the contrastive task. During fine-tuning, following AimTS \cite{AimTS}, we use a standard classifier on the temporal pathway representations to adapt them to the downstream task. This ensures a fair comparison with baselines that also use single-modality features .

\section{Experiments}
\subsection{Experimental Setup}
\textbf{Datasets.}
We conduct experiments on datasets spanning three clinical diagnostic scenarios.
(1) \textbf{Alzheimer’s Disease}: APAVA \cite{escudero2006analysis} and ADFTD \cite{miltiadous2023dataset} are two EEG datasets
for Alzheimer’s disease classification.
(2) \textbf{Epilepsy}: TUSZ v1.5.2 \cite{shah2018temple} is a large-scale EEG dataset for epilepsy. 
TUSZ (2-Classes) provides a coarse-grained seizure/non-seizure setting, while TUSZ (4-Classes) provides a fine-grained four-class seizure taxonomy.
(3) \textbf{Cardiac Disease}: PTB \cite{physiobank2000physionet} and PTB-XL \cite{wagner2020ptb} are two large-scale ECG databases for cardiac diagnosis. PTB-XL (4-Classes) corresponds to a coarse-grained 4-class setting, while PTB-XL (5-Classes) corresponds to a fine-grained 5-class setting. Table \ref{tab:dat} shows data details.
%while \textbf{Appendix \ref{dat_app} provides more information, including the disease classes, dataset URLs, associated clinical text information, train–validation–test splits, and pre-processing procedures.}

\begin{table}[tb] % 双栏中的跨栏

\caption{Datasets Statistics.} %注意要求是在上方，还是表格下方
\label{tab:dat} %引用
%\vspace{-3mm} %表格和其上排版的距离
\centering % 正文居中

\vspace{-3mm} % 表格内容和标题的距离 
    
    \resizebox{\columnwidth}{!}{ % 盒子，表格占页面宽度
    %\scriptsize
    %\footnotesize
        \begin{threeparttable} %三栏式，可以在下面加标注

\begin{tabular}{>{\bfseries}l|c|c|c|c}
\midrule \midrule
\textbf{Datasets} & \textbf{Total Samples} & \textbf{Classes} & \textbf{Channels} & \textbf{Steps} \\ 
\midrule
APAVA (2-Classes)   & 5,967  & 2 & 16 & 256  \\ 
\midrule
ADFTD (3-Classes)   & 69,752 & 3 & 19 & 256  \\ 
\midrule
TUSZ (2-Classes)    & 22,040 & 2 & 19 & 6,000 \\ 
\midrule
TUSZ (4-Classes)    & 2,891  & 4 & 19 & 6,000 \\ 
\midrule
PTB (2-Classes)     & 64,356 & 2 & 15 & 300  \\ 
\midrule
PTB-XL (4-Classes)  & 17,110 & 4 & 12 & 1,000 \\ 
\midrule
PTB-XL (5-Classes)  & 17,110 & 5 & 12 & 1,000 \\
\midrule \midrule
\end{tabular}
            
            %\begin{tablenotes} 
            %\item[1] Note: ...
            %\end{tablenotes}  % 注释，如果标题在下方，这个无法使用
        
        \end{threeparttable} % 三栏式，可以在下面加标注
        
        } % 盒子
\vspace{-3mm} %表格和其下排版的距离
\end{table} % 双栏中的跨栏
%\end{table} % 单栏
%\end{wraptable} %文字围绕表格

% \text{\small ±0.15} 
% \small \scriptsize
% - &  }&  -  } -} }}

% 只保留F1
%\begin{wraptable}{r}{0.6\textwidth} %文字围绕表格
%\begin{table}[htb] % 单栏
\begin{table*}[htb] % 双栏中的跨栏
%\vspace{-4mm} %表格和其上排版的距离
\caption{\textbf{Supervised Learning}. 
%Full results of seven datasets see \textbf{Appendix \ref{sup_app}}. 
\textcolor{red}{Red}: best, \textcolor{blue}{Blue}: second best. 
Modalities are denoted as: \textbf{T} for time-series signals, \textbf{V} for visual representations, and \textbf{L} for language.}%注意要求是在上方，还是表格下方
\centering % 正文居中

\vspace{-3mm} % 表格内容和标题的距离
    
    \resizebox{\textwidth}{!}{ % 盒子，表格占页面宽度
    %\scriptsize
    \footnotesize
        \begin{threeparttable} %三栏式，可以在下面加标注
%\begin{tabular}{l|cc|cc|cc|cc|cc}
\begin{tabular}{>{\bfseries}l|cc|cc|cc|cc|cc}
%\begin{tabular}{lrrrrrrr}
%\toprule 
\midrule \midrule

% \multicolumn{1}{l|}{\multirow{2}{*}{\begin{tabular}[c]{@{}l@{}}Methods\\ (Modality)\end{tabular}}} &
%   \multicolumn{2}{c|}{APAVA (2-Classes)} &
%   \multicolumn{2}{c|}{ADFTD (3-Classes)} &
%   \multicolumn{2}{c|}{TUSZ (4-Classes)} &
%   \multicolumn{2}{c|}{PTB-XL (4-Classes) } &
%   \multicolumn{2}{c}{PTB (2-Classes)} \\
% \multicolumn{1}{l|}{} &
%   F1 &
%   \multicolumn{1}{c|}{Acc.} &
%   F1 &
%   \multicolumn{1}{c|}{Acc.} &
%   F1 &
%   \multicolumn{1}{c|}{Acc.} &
%   F1 &
%   \multicolumn{1}{c|}{Acc.} &
%   F1 & Acc.\\ \midrule

% 2. 第一行：手动为表头文字加粗
        \multicolumn{1}{l|}{\multirow{2}{*}{\begin{tabular}[c]{@{}l@{}}\textbf{Methods}\\ \textbf{(Modality)}\end{tabular}}} &
        \multicolumn{2}{c|}{\textbf{APAVA (2-Classes)}} &
        \multicolumn{2}{c|}{\textbf{ADFTD (3-Classes)}} &
        \multicolumn{2}{c|}{\textbf{TUSZ (4-Classes)}} &
        \multicolumn{2}{c|}{\textbf{PTB-XL (4-Classes)}} &
        \multicolumn{2}{c}{\textbf{PTB (2-Classes)}} \\
        
        \multicolumn{1}{l|}{} &
        \textbf{F1} & \multicolumn{1}{c|}{\textbf{Acc.}} &
        \textbf{F1} & \multicolumn{1}{c|}{\textbf{Acc.}} &
        \textbf{F1} & \multicolumn{1}{c|}{\textbf{Acc.}} &
        \textbf{F1} & \multicolumn{1}{c|}{\textbf{Acc.}} &
        \textbf{F1} & \textbf{Acc.} \\ \midrule
        
Dlinear (T) & 56.19\text{\scriptsize $\pm$1.27} & 65.48\text{\scriptsize $\pm$0.33} & 39.58\text{\scriptsize $\pm$0.84} & 46.96\text{\scriptsize $\pm$2.11} & 70.70\text{\scriptsize $\pm$0.36} & 80.14\text{\scriptsize $\pm$0.88} & 33.71\text{\scriptsize $\pm$2.79} & 57.01\text{\scriptsize $\pm$2.68} & 62.78\text{\scriptsize $\pm$0.69} & 74.42\text{\scriptsize $\pm$0.51} \\ 
MedGNN (T) & 80.78\text{\scriptsize $\pm$2.95} & 81.91\text{\scriptsize $\pm$1.81} & 45.42\text{\scriptsize $\pm$1.07} & 47.71\text{\scriptsize $\pm$2.62} & 83.06\text{\scriptsize $\pm$0.76} & 90.16\text{\scriptsize $\pm$2.46} & 66.25\text{\scriptsize $\pm$2.24} & 77.21\text{\scriptsize $\pm$0.74} & 80.58\text{\scriptsize $\pm$2.35} & 84.36\text{\scriptsize $\pm$2.96} \\ 
\midrule
    MultiRocket (T) & 58.81\text{\scriptsize $\pm$1.63} & 58.84\text{\scriptsize $\pm$1.50} & 36.59\text{\scriptsize $\pm$2.83} & 51.32\text{\scriptsize $\pm$2.37} & 75.55\text{\scriptsize $\pm$2.46} & 84.28\text{\scriptsize $\pm$1.08} & 50.93\text{\scriptsize $\pm$2.19} & 63.35\text{\scriptsize $\pm$0.82} & 64.11\text{\scriptsize $\pm$1.06} & 74.78\text{\scriptsize $\pm$2.40} \\
    ResNet (T) & 69.67\text{\scriptsize $\pm$2.35} & 72.26\text{\scriptsize $\pm$1.76} & 40.41\text{\scriptsize $\pm$1.05} & 45.91\text{\scriptsize $\pm$2.30} & 78.35\text{\scriptsize $\pm$2.93} & 86.36\text{\scriptsize $\pm$2.58} & 63.79\text{\scriptsize $\pm$0.64} & 77.25\text{\scriptsize $\pm$0.49} & 68.14\text{\scriptsize $\pm$2.49} & 75.97\text{\scriptsize $\pm$2.87} \\
    InceptionTime (T) & 78.36\text{\scriptsize $\pm$2.90} & 80.92\text{\scriptsize $\pm$2.02} & 45.79\text{\scriptsize $\pm$1.35} & 53.19\text{\scriptsize $\pm$1.95} & 82.95\text{\scriptsize $\pm$2.26} & 88.77\text{\scriptsize $\pm$0.43} & 68.71\text{\scriptsize $\pm$2.84} & 77.21\text{\scriptsize $\pm$2.61} & 78.91\text{\scriptsize $\pm$2.37} & 83.56\text{\scriptsize $\pm$2.29} \\
    TimesNet (T) & 73.12\text{\scriptsize $\pm$2.87} & 75.75\text{\scriptsize $\pm$1.74} & 46.09\text{\scriptsize $\pm$0.67} & 50.43\text{\scriptsize $\pm$2.39} & 83.86\text{\scriptsize $\pm$2.03} & 88.43\text{\scriptsize $\pm$2.66} & 64.13\text{\scriptsize $\pm$2.94} & 75.35\text{\scriptsize $\pm$2.00} & 73.11\text{\scriptsize $\pm$2.32} & 79.23\text{\scriptsize $\pm$2.83} \\ \midrule
    PatchTST (T) & 61.93\text{\scriptsize $\pm$2.94} & 67.99\text{\scriptsize $\pm$2.05} & 40.41\text{\scriptsize $\pm$1.09} & 43.24\text{\scriptsize $\pm$1.81} & 81.03\text{\scriptsize $\pm$2.32} & 81.87\text{\scriptsize $\pm$1.31} & 57.97\text{\scriptsize $\pm$1.41} & 74.14\text{\scriptsize $\pm$2.49} & 72.77\text{\scriptsize $\pm$1.82} & 80.27\text{\scriptsize $\pm$0.78} \\
    iTransformer (T) & 74.31\text{\scriptsize $\pm$1.17} & 76.11\text{\scriptsize $\pm$2.13} & 41.57\text{\scriptsize $\pm$2.73} & 45.41\text{\scriptsize $\pm$1.92} & 82.27\text{\scriptsize $\pm$0.66} & 87.56\text{\scriptsize $\pm$1.06} & 61.41\text{\scriptsize $\pm$2.28} & 74.01\text{\scriptsize $\pm$0.39} & 75.52\text{\scriptsize $\pm$2.90} & 82.01\text{\scriptsize $\pm$1.40} \\
    Medformer (T) & 72.74\text{\scriptsize $\pm$0.61} & 76.59\text{\scriptsize $\pm$2.81} & 46.23\text{\scriptsize $\pm$2.84} & 53.79\text{\scriptsize $\pm$1.28} & 84.27\text{\scriptsize $\pm$2.57} & 88.77\text{\scriptsize $\pm$2.48} & 62.26\text{\scriptsize $\pm$2.11} & 76.88\text{\scriptsize $\pm$2.09} & 77.37\text{\scriptsize $\pm$2.36} & 82.41\text{\scriptsize $\pm$2.98} \\ 
    ViTST (V) & 80.93\text{\scriptsize $\pm$0.78} & 81.97\text{\scriptsize $\pm$2.33} & 41.57\text{\scriptsize $\pm$2.89} & 45.41\text{\scriptsize $\pm$1.39} & 82.74\text{\scriptsize $\pm$2.05} & 88.08\text{\scriptsize $\pm$1.04} & 65.22\text{\scriptsize $\pm$2.99} & 75.58\text{\scriptsize $\pm$2.64} & 82.62\text{\scriptsize $\pm$2.31} & 85.85\text{\scriptsize $\pm$2.48} \\
        MedViA (TV) & \textcolor{blue}{81.26\text{\scriptsize $\pm$2.96}} & \textcolor{blue}{83.44\text{\scriptsize $\pm$1.82}} & 47.85\text{\scriptsize $\pm$1.08} & 51.71\text{\scriptsize $\pm$2.65} & \textcolor{blue}{86.51\text{\scriptsize $\pm$0.78}} & 89.12\text{\scriptsize $\pm$2.48} & 67.89\text{\scriptsize $\pm$2.25} & 79.75\text{\scriptsize $\pm$0.76} & 83.93\text{\scriptsize $\pm$2.36} & 86.72\text{\scriptsize $\pm$2.99} \\ \midrule
    GPT4TS (T) & 77.92\text{\scriptsize $\pm$0.94} & 80.92\text{\scriptsize $\pm$1.83} & 43.72\text{\scriptsize $\pm$2.54} & 51.51\text{\scriptsize $\pm$2.44} & 83.35\text{\scriptsize $\pm$1.84} & \textcolor{red}{91.19\text{\scriptsize $\pm$1.33}} & 63.11\text{\scriptsize $\pm$2.86} & 74.98\text{\scriptsize $\pm$0.79} & 76.82\text{\scriptsize $\pm$2.33} & 82.32\text{\scriptsize $\pm$1.64} \\
    MedTsLLM (TL) & 78.24\text{\scriptsize $\pm$2.99} & 78.76\text{\scriptsize $\pm$1.87} & 45.83\text{\scriptsize $\pm$1.41} & 52.59\text{\scriptsize $\pm$2.63} & 83.95\text{\scriptsize $\pm$2.20} & 89.29\text{\scriptsize $\pm$2.49} & 64.21\text{\scriptsize $\pm$2.87} & 76.41\text{\scriptsize $\pm$0.77} & 75.66\text{\scriptsize $\pm$2.36} & 81.31\text{\scriptsize $\pm$2.97} \\
    MedualTime (TL) & 79.85\text{\scriptsize $\pm$2.94} & 81.62\text{\scriptsize $\pm$1.83} & 46.95\text{\scriptsize $\pm$1.11} & 50.77\text{\scriptsize $\pm$2.64} & 85.37\text{\scriptsize $\pm$0.83} & 90.51\text{\scriptsize $\pm$2.47} & \textcolor{blue}{72.53\text{\scriptsize $\pm$2.27}} & \textcolor{blue}{80.58\text{\scriptsize $\pm$0.75}} & 77.68\text{\scriptsize $\pm$2.38} & 81.78\text{\scriptsize $\pm$2.98} \\
    TimeVLM (TVL) & 79.93\text{\scriptsize $\pm$2.98} & 82.39\text{\scriptsize $\pm$1.84} & \textcolor{blue}{48.35\text{\scriptsize $\pm$1.13}} & \textcolor{blue}{53.84\text{\scriptsize $\pm$2.66}} & 85.72\text{\scriptsize $\pm$0.81} & \textcolor{blue}{90.85\text{\scriptsize $\pm$2.49}} & 70.29\text{\scriptsize $\pm$2.28} & 80.31\text{\scriptsize $\pm$0.78} & \textcolor{blue}{84.71\text{\scriptsize $\pm$2.37}} & \textcolor{blue}{87.48\text{\scriptsize $\pm$2.99}} \\ \midrule
    \textbf{MedTVL (TVL)} & \textcolor{red}{85.52\text{\scriptsize $\pm$2.97}} & \textcolor{red}{85.88\text{\scriptsize $\pm$1.85}} & \textcolor{red}{51.27\text{\scriptsize $\pm$1.15}} & \textcolor{red}{54.12\text{\scriptsize $\pm$2.67}} & \textcolor{red}{89.26\text{\scriptsize $\pm$0.82}} & \textcolor{red}{91.19\text{\scriptsize $\pm$2.50}} & \textcolor{red}{77.93\text{\scriptsize $\pm$2.29}} & \textcolor{red}{84.43\text{\scriptsize $\pm$0.79}} & \textcolor{red}{88.22\text{\scriptsize $\pm$2.39}} & \textcolor{red}{90.04\text{\scriptsize $\pm$2.99}}

% \end{tabular}
% \end{table}
            \\ % 这里要换行才能添加bottom rule
            %\bottomrule 
            \midrule \midrule
            \end{tabular} % 表尾
            
            %\begin{tablenotes} 
            %\item[1] Note: ...
            %\end{tablenotes}  % 注释，如果标题在下方，这个无法使用
        
        \end{threeparttable} % 三栏式，可以在下面加标注
        
        } % 盒子

\label{tab:sup} %引用
\vspace{-3mm} %表格和其下排版的距离
\end{table*} % 双栏中的跨栏
%\end{table} % 单栏
%\end{wraptable} %文字围绕表格

\textbf{Baselines.}
For a comprehensive comparison, we choose representative baselines from diverse architectures and modalities as follows: 
\textbf{(1) MLP}:  DLinear \cite{zeng2023dlinear}; 
\textbf{(2) GNN}: MedGNN \cite{MedGNN}; 
\textbf{(3) CNN}: ResNet \cite{he2016deep}, MultiRocket \cite{tan2022multirocket}, InceptionTime \cite{InceptionTime} and TimesNet \cite{wu2022timesnet}; \textbf{(4) Transformer}: PatchTST \cite{nie2023time}, Medformer \cite{Medformer}, iTransformer \cite{LiuHZWWML24}, ViTST \cite{ViTST} and MedViA \cite{MedViA}; \textbf{(5) LLM/VLM}: GPT4TS \cite{GPT4TS}, MedTsLLM \cite{MedTsLLM}, MedualTime \cite{MedualTime}, and Time-VLM \cite{Time-VLM}.
Our contrastive learning baselines include T-Loss \cite{T-Loss}, TS-TCC \cite{TS-TCC}, TS2Vec \cite{Ts2vec}, InfoTS \cite{InfoTS}, TimesURL \cite{TimesURL}, and AimTS \cite{AimTS}.
%\textbf{More baselines information are provided in the Appendix \ref{bas_app}}.

\textbf{Implementation Details.}
Following \cite{MedViA},  we report macro-averaged F1, AUROC, AUPRC, accuracy, precision, and recall. We prioritize F1 score due to data imbalance while other metrics deferred to the \textbf{Appendix}.
For a fair comparison and to maintain consistent experimental conditions, all models are integrated into a unified implementation framework. All models are trained with a batch size of 32 using the AdamW optimizer and a cosine scheduler for up to 50 epochs, supplemented by an early stopping patience of 7. To ensure statistical reliability, we report the mean and standard deviation over 6 independent runs (random seeds 42-47) on fixed dataset splits. The optimal model is selected based on the highest validation F1-score and subsequently evaluated on the test set.
For MedTVL, the Swin-Base architecture (patch size 4, window size 7, image size 224 $\times$ 224) pre-trained on ImageNet-21k serves as the visual backbone. We fine-tune only the first two encoder stages $
S=(0,1)$ to balance efficacy and efficiency, refining low-level medical textures while preserving high-level pre-trained priors. InceptionTime is adopted as the temporal backbone following the original configuration \cite{InceptionTime}, featuring a depth of 6, multi-scale kernels $[39, 19, 9]$, and filter channels $nf=32$. ClinicalBERT \cite{wang2023optimized} is employed as a frozen language model. As to MoE module, we set expert number $O=4$ and top $K=2$ routed experts to handle the pathway heterogeneity.
The dimension of hidden embedding  is  $128$ and output embedding is $D=64$. 
We optimize baselines by tuning critical hyper-parameters.
All experiments are using PyTorch on NVIDIA A$6000$ ($48$GB) GPU. 
%\textbf{Baseline configurations are in Appendix \ref{imp_app} and image construction  are in Appendix \ref{grid_app}}.

\subsection{Supervised Learning}
Key findings in Table \ref{tab:sup} include:
(1) Classical linear or shallow models such as DLinear and MultiRocket  usually underperform across all datasets.
(2) Among CNN-based methods, InceptionTime usually achieves the best or second-best performance, particularly on ECG signals.
(3) Multi-modal approaches generally outperform single-modality baselines across most datasets, indicating that incorporating cross-modal information improves robustness.
(4) MedTVL consistently outperforms competing methods across datasets and metrics. Notably, it achieves an improvement of approximately \textbf{3\%–5\%} in average F1 over the second-best models, highlighting the effectiveness of the proposed tri-modal framework and the text-assisted dual-pathway design.

\begin{figure*}[h]
%\begin{figure*}[htb] % 双栏中的跨栏
%\begin{wrapfigure}{r}{0.6\textwidth}  % 'r' 表示右对齐，0.6\textwidth 表示图像将占据页面宽度的 60%。
%\vspace{-2mm}
\centering

    % 单个图片, 双栏中的单栏图width=0.49，双栏中的跨栏图 width=0.98
    %\includegraphics[width=0.49\textwidth, height=0.35\textheight]{./Figures/pdfs/AHGCSP.pdf}
    \includegraphics[width=0.99\textwidth]{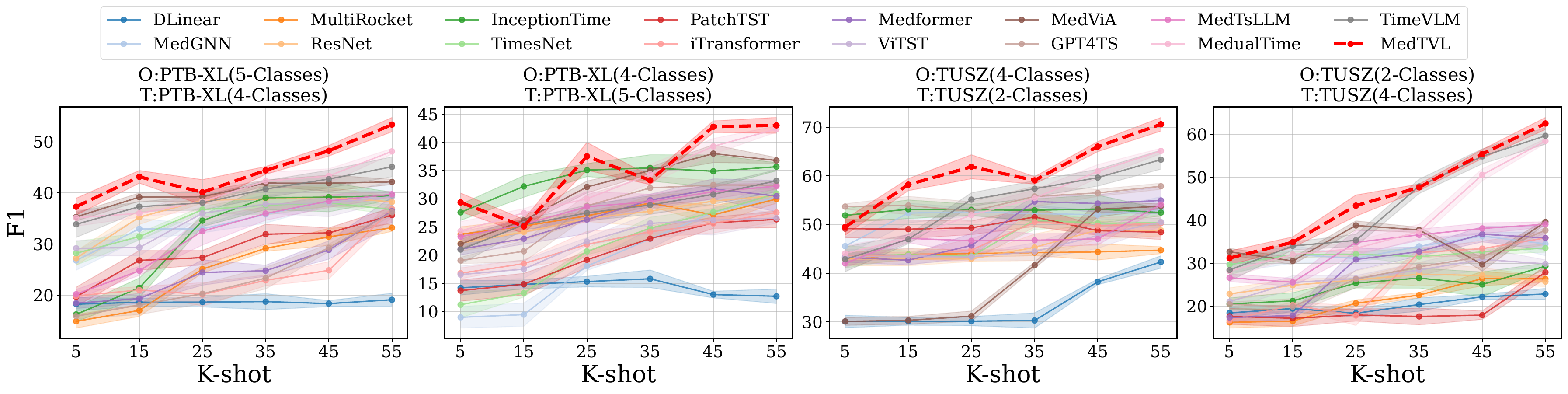}
      
    %\vspace{-3mm} % 图片与标题的排版距离
    \caption{
    Few-shot  performance in F1 score under different shot settings. Other metrics are in \textbf{Appendix \ref{abl_app}}.
    \textit{O} denotes the source domain and \textit{T} denotes the target domain.}
    
    %  The red dashed line is MedSpaformer.
\label{fig:fewshot_main}
%\vspace{-3mm} % 图片标题的下方排版距离
\end{figure*}
%\end{figure*} % 双栏中的跨栏
%\end{wrapfigure}
\subsection{Few-shot Learning}
We conduct few-shot learning to alleviate the label scarcity challenge by transferring knowledge from a label-rich source domain to a target domain with restricted annotations. Specifically, models are first pre-trained on the source dataset and subsequently fine-tuned on the target dataset under various $k$-shot settings ($k \in \{5, 15, 25, 35, 45, 55\}$). Due to the rigid input dimension requirements of our baselines, we select source-target dataset pairs with identical timesteps and channels, namely PTB-XL (4-Classes) and PTB-XL (5-Classes), TUSZ (2-Classes) and TUSZ (4-Classes). During the fine-tuning phase, the pre-trained backbones remain frozen, and only a newly initialized classification head is optimized.
As shown in Figure \ref{fig:fewshot_main}: (1) Model performance generally scales positively with the number of shots, with our model achieving state-of-the-art results in nearly all settings, demonstrating robust cross-dataset transferability. (2) Notably, our model exhibits sustained performance gains as the number of shots increases, suggesting that it can effectively internalize additional information without reaching an early plateau. (3) Transferring from fine-grained to coarse-grained label sets  yields superior results than the reverse. This suggests that fine-grained pre-training compels the model to learn more nuanced, discriminative features for handling downstream tasks.

%\begin{wraptable}{r}{0.6\textwidth} %文字围绕表格
\begin{table}[tb] % 单栏
%\begin{table*}[htb] % 双栏中的跨栏
\vspace{-4mm} %表格和其上排版的距离
\caption{
Contrastive learning via linear probing on 100\% labeled set in F1 score. 
%Full results are in \textbf{Appendix \ref{unsup_app}}.
}
\centering % 正文居中
%\caption{Caption} %注意要求是在上方，还是表格下方
\vspace{-3mm} % 表格内容和标题的距离
    
    \resizebox{\columnwidth}{!}{ % 盒子，表格占页面宽度
        \begin{threeparttable} %三栏式，可以在下面加标注

    % 表格所有内容字体统一设为11pt，行间距13pt（确保清晰）
    \begin{tabular}{>{\bfseries\fontsize{12}{14}\selectfont}l|>{\fontsize{12}{14}\selectfont}c|>{\fontsize{12}{14}\selectfont}c|>{\fontsize{12}{14}\selectfont}c|>{\fontsize{12}{14}\selectfont}c|>{\fontsize{12}{14}\selectfont}c}
      \midrule \midrule
      % 表头字体同步放大
      \multicolumn{1}{l|}{\textbf{\begin{tabular}[c]{@{}l@{}}\fontsize{12}{14}\selectfont Methods\\ \fontsize{12}{14}\selectfont (Modality)\end{tabular}}} &
      \textbf{\makecell{\fontsize{12}{14}\selectfont APAVA\\\fontsize{12}{14}\selectfont (2-Classes)}} &
      \textbf{\makecell{\fontsize{12}{14}\selectfont ADFTD\\\fontsize{12}{14}\selectfont (3-Classes)}} &
      \textbf{\makecell{\fontsize{12}{14}\selectfont TUSZ\\\fontsize{12}{14}\selectfont (4-Classes)}} &
      \textbf{\makecell{\fontsize{12}{14}\selectfont PTB-XL\\\fontsize{12}{14}\selectfont (4-Classes)}} &
      \textbf{\makecell{\fontsize{12}{14}\selectfont PTB\\\fontsize{12}{14}\selectfont (2-Classes)}} \\ \midrule
      
      % 数值和方差字体均放大（方差用\small而非\scriptsize，避免过小）
      T-Loss (T)     & 49.44\text{\small $\pm$1.12} & 36.72\text{\small $\pm$3.85} & 64.03\text{\small $\pm$1.34} & 33.87\text{\small $\pm$3.62} & 50.88\text{\small $\pm$1.05} \\
      TSTCC (T)      & 57.88\text{\small $\pm$1.07} & 37.55\text{\small $\pm$3.68} & 71.35\text{\small $\pm$1.29} & 48.15\text{\small $\pm$3.47} & 63.51\text{\small $\pm$1.13} \\
      TS2Vec (T)     & 62.33\text{\small $\pm$1.02} & 39.26\text{\small $\pm$3.53} & 73.58\text{\small $\pm$1.24} & 49.01\text{\small $\pm$3.32} & 64.65\text{\small $\pm$1.18} \\
      InfoTS (T)     & 64.73\text{\small $\pm$0.97} & 43.72\text{\small $\pm$3.37} & 72.22\text{\small $\pm$1.19} & 53.38\text{\small $\pm$3.19} & 64.11\text{\small $\pm$1.23} \\
      TimesURL (T)   & 62.86\text{\small $\pm$1.05} & 43.40\text{\small $\pm$3.41} & \textcolor{blue}{75.91\text{\small $\pm$1.21}} & 59.29\text{\small $\pm$3.25} & 66.41\text{\small $\pm$1.16} \\ \midrule
      AimTS (TV)      & \textcolor{blue}{72.65\text{\small $\pm$0.91}} & \textcolor{blue}{44.16\text{\small $\pm$3.23}} & 74.77\text{\small $\pm$1.14} & \textcolor{blue}{60.81\text{\small $\pm$3.08}} & \textcolor{blue}{71.32\text{\small $\pm$1.19}} \\ \midrule
      MedTVL (TVL)     & \textcolor{red}{76.20\text{\small $\pm$0.86}} & \textcolor{red}{45.16\text{\small $\pm$3.11}} & \textcolor{red}{79.18\text{\small $\pm$1.09}} & \textcolor{red}{67.53\text{\small $\pm$2.97}} & \textcolor{red}{74.56\text{\small $\pm$1.14}}

            \\ % 这里要换行才能添加bottom rule
            \midrule \midrule
            \end{tabular} % 表尾
            
            %\begin{tablenotes} 
            %\item[1] Note: ...
            %\end{tablenotes}  % 注释，如果标题在下方，这个无法使用
        
        \end{threeparttable} % 三栏式，可以在下面加标注
        
        } % 盒子

%\vspace{-3mm} % 表格内容和标题的距离
\label{tab:unsup} %引用
\vspace{-3mm} %表格和其下排版的距离
%\end{table*} % 双栏中的跨栏
\end{table} % 单栏
%\end{wraptable} %文字围绕表格
\begin{figure*}[h]
%\begin{figure*}[htb] % 双栏中的跨栏
%\begin{wrapfigure}{r}{0.6\textwidth}  % 'r' 表示右对齐，0.6\textwidth 表示图像将占据页面宽度的 60%。
%\vspace{-3mm}
\centering

    % 单个图片, 双栏中的单栏图width=0.49，双栏中的跨栏图 width=0.98
    %\includegraphics[width=0.49\textwidth, height=0.35\textheight]{./Figures/pdfs/AHGCSP.pdf}
    \includegraphics[width=0.99\textwidth]{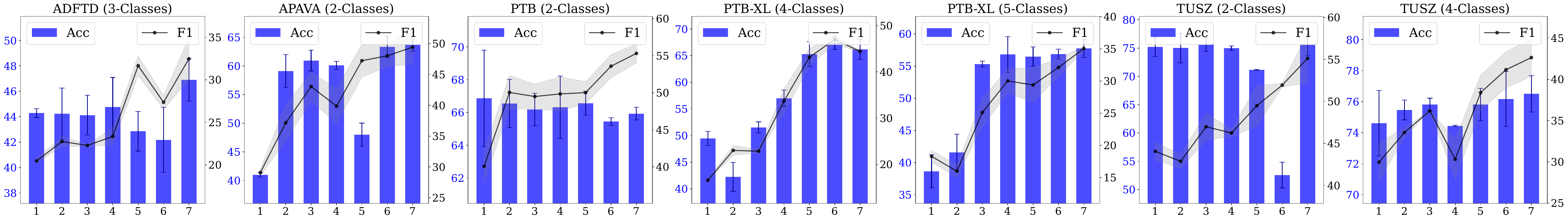}
      
    %\vspace{-4mm} % 图片与标题的排版距离
    \caption{
    Contrastive learning via linear probing on 10\% labeled set. 1=T-Loss, 2=TSTCC, 3=TS2Vec, 4=InfoTS, 5=TimesURL, 6=AimTS, 7=MedTVL.}
    
    %  The red dashed line is MedSpaformer.
\label{fig:con_10_app}
\vspace{-2mm} % 图片标题的下方排版距离
\end{figure*}
%\end{figure*} % 双栏中的跨栏
%\end{wrapfigure}
\begin{figure*}[h]
%\begin{figure*}[htb] % 双栏中的跨栏
%\begin{wrapfigure}{r}{0.6\textwidth}  % 'r' 表示右对齐，0.6\textwidth 表示图像将占据页面宽度的 60%。
%\vspace{-3mm}
\centering

    % 单个图片, 双栏中的单栏图width=0.49，双栏中的跨栏图 width=0.98
    %\includegraphics[width=0.49\textwidth, height=0.35\textheight]{./Figures/pdfs/AHGCSP.pdf}
    \includegraphics[width=0.99\textwidth]{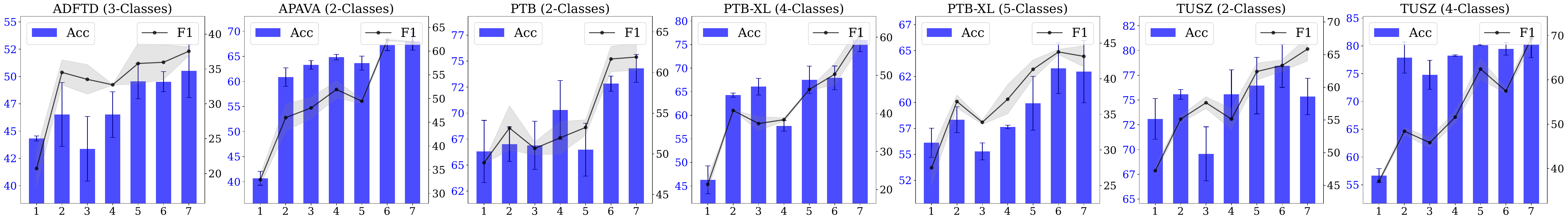}
      
    %\vspace{-4mm} % 图片与标题的排版距离
    \caption{
    Contrastive learning via linear probing on 50\% labeled set. 1=T-Loss, 2=TSTCC, 3=TS2Vec, 4=InfoTS, 5=TimesURL, 6=AimTS, 7=MedTVL.}
    
    %  The red dashed line is MedSpaformer.
\label{fig:con_50_app}
%\vspace{-3mm} % 图片标题的下方排版距离
\end{figure*}
%\end{figure*} % 双栏中的跨栏
%\end{wrapfigure}
\subsection{Contrastive Learning}
We first pre-train our framework to produce unsupervised embeddings, and then evaluate their quality and label efficiency by training a linear classifier under both full-resource ($100\%$ labeled training data) and data-sparse ($5\%$ and $50\%$ labeled training data) settings, while keeping the validation and test sets fixed across all experiments. The results for the 100\% data setting are  summarized in Table \ref{tab:unsup} while Figure \ref{fig:con_10_app} and \ref{fig:con_50_app} show results under 10\% and 50\% labeled data settings.
Table \ref{tab:unsup} shows that multimodal models consistently outperform time-only baselines, highlighting the synergy between visual and temporal signal. While AimTS shows strong transferability, our MedTVL achieves the best results across all metrics. Notably, MedTVL surpasses the strongest baseline by 6.72\% in PTB-XL (4-Classes). This consistent lead demonstrates that tri-modal fusion provides richer information density, enabling the backbone to learn more discriminative  representations.

\subsection{More Experiments}
\textbf{(a) Ablation Study.}
To evaluate the impact of critical modules, we compare five ablation variants: "w/o MoE (MLP)" replaces MoE with a standard MLP; "w/o MoE (Gated Fusion)" replaces MoE with a gated fusion mechanism \cite{Time-VLM}; "w/o Textual Guidance" removes the Adaptive Textual Guidance module; "w/o Visual Pathway" and "w/o Temporal Pathway" respectively exclude the visual and temporal branches.
Table \ref{tab:abl} shows that the dual pathways emerge as the most fundamental component.
The removal of the temporal pathway yields the steepest F1 decline on PTB-XL (about 10.4\%), while excluding the visual pathway results in an average F1 drop of about 5\% on TUSZ and APAVA.
This suggests that different medical datasets rely on distinct modal features.
When the MoE module is removed, gated fusion outperforms MLP-based fusion, likely because it can assign different weights to modal features. However, these weights are shared across samples rather than dynamically adapted to individual instances.
The MoE module and textual guidance further enhance the model, yielding approximately 3\% and 2\% F1 improvements, respectively.
These results validate the synergy of our tri-modal design and the MoE-based fusion in capturing sample-wise physiological patterns.

\begin{table}[htb] % 双栏中的跨栏

\caption{Ablation study in F1 score on three datasets. 
%Full results are in \textbf{Appendix \ref{abl_app}}.
} %注意要求是在上方，还是表格下方
\label{tab:abl} %引用
\vspace{-3mm} %表格和其上排版的距离
\centering % 正文居中
    
    \resizebox{\columnwidth}{!}{ % 盒子，表格占页面宽度
    %\scriptsize
    %\footnotesize
        \begin{threeparttable} %三栏式，可以在下面加标注
        
\begin{tabular}{>{\bfseries}l|c|c|c}
    \midrule \midrule
    \textbf{Variants} &
    \textbf{\makecell{APAVA\\(2-Classes)}} &
    \textbf{\makecell{TUSZ\\(4-Classes)}} &
    \textbf{\makecell{PTB-XL\\(4-Classes)}} \\
    \midrule
    w/o MoE (MLP)        & 82.60\text{\scriptsize $\pm$1.23} & 86.67\text{\scriptsize $\pm$2.58} & 74.18\text{\scriptsize $\pm$3.12} \\
    w/o MoE (Gated Fusion)  & 83.33\text{\scriptsize $\pm$1.87} & \textcolor{blue}{87.28\text{\scriptsize $\pm$2.45}} & \textcolor{blue}{75.77\text{\scriptsize $\pm$2.91}} \\
    w/o Textual Guidance  & \textcolor{blue}{83.35\text{\scriptsize $\pm$2.11}} & 86.56\text{\scriptsize $\pm$3.87} & 75.64\text{\scriptsize $\pm$1.45} \\
    w/o Visual Pathway    & 80.43\text{\scriptsize $\pm$1.78} & 84.27\text{\scriptsize $\pm$2.98} & 70.58\text{\scriptsize $\pm$3.45} \\
    w/o Temporal Pathway  & 81.11\text{\scriptsize $\pm$2.56} & 85.19\text{\scriptsize $\pm$1.23} & 67.53\text{\scriptsize $\pm$0.87} \\ \midrule
    MedTVL                & \textcolor{red}{85.52\text{\scriptsize $\pm$2.97}} & \textcolor{red}{89.26\text{\scriptsize $\pm$0.82}} & \textcolor{red}{77.93\text{\scriptsize $\pm$2.29}} 
    
            \\ % 这里要换行才能添加bottom rule
            %\bottomrule 
            \midrule \midrule
            \end{tabular} % 表尾
            
            %\begin{tablenotes} 
            %\item[1] Note: ...
            %\end{tablenotes}  % 注释，如果标题在下方，这个无法使用
        
        \end{threeparttable} % 三栏式，可以在下面加标注
        
        } % 盒子
\vspace{-3mm} %表格和其下排版的距离
\end{table} % 双栏中的跨栏
%\end{table} % 单栏
%\end{wraptable} %文字围绕表格

%\begin{wrapfigure}{r}{0.45\textwidth}  % "r" 表示图像在右侧，0.5\textwidth 表示图像宽度为页面宽度的一半
% \begin{figure}[htb]
\begin{figure}[htb]
  \vspace{-2mm}
  \begin{center}
    \includegraphics[width=0.45\textwidth]{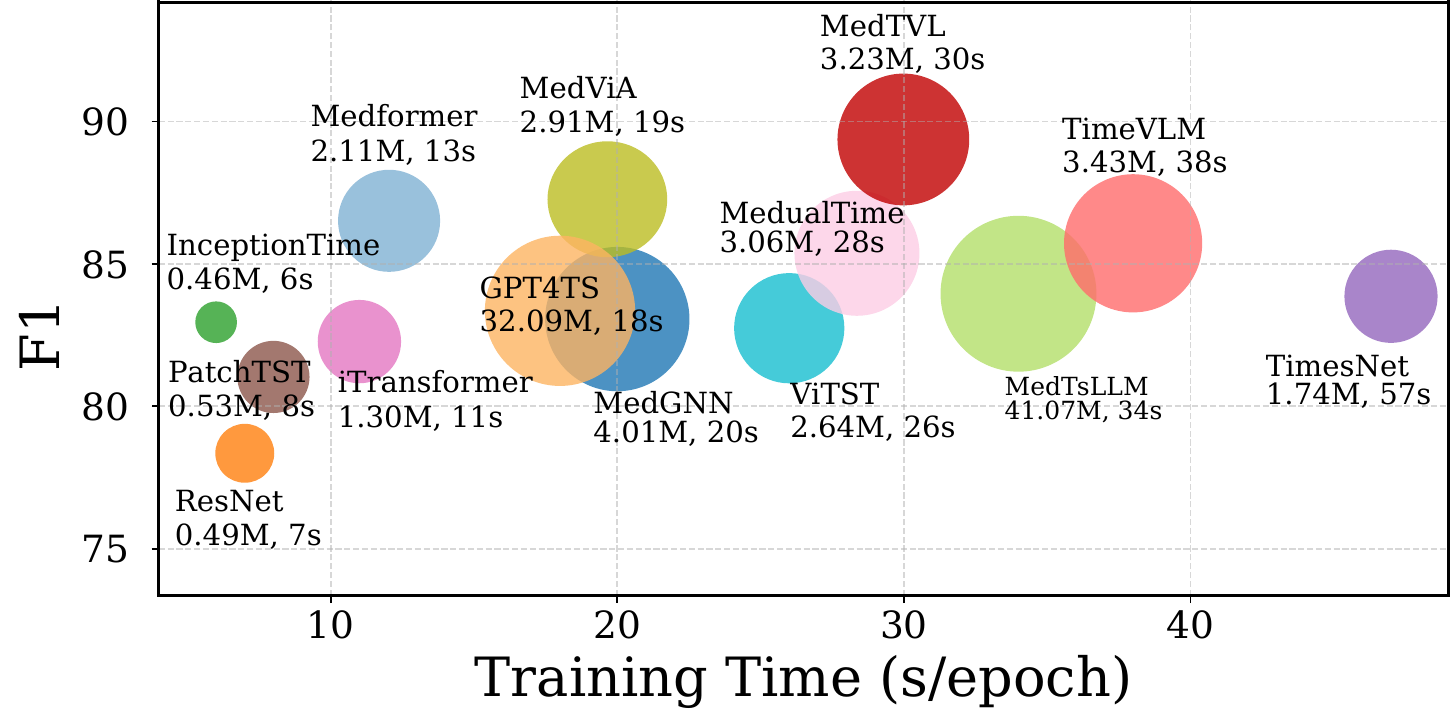} % 这里调整图像的宽度，确保它在wrapfigure中居中显示
  \end{center}
  \vspace{-3mm}
  \caption{Efficiency comparison on TUSZ (4-Classes), where \textbf{M} denotes millions of trainable parameters.}
  %The bubble size represents the the model parameter size.
  % The dotted area denotes the amount of trainable parameters of different models.
  % Detection and classification tasks can be distinguished by marker type.}
    \label{fig:efficiency}
    \vspace{-3mm} 
%\end{wrapfigure}
\end{figure}
\textbf{(b) Efficiency Analysis.}
In Figure \ref{fig:efficiency}, we compare the efficiency of MedTVL with representative baselines on the TUSZ (4- Classes), considering training time per epoch, F1 score, and the number of trainable parameters.
(1) Among all methods, InceptionTime is the most lightweight and fastest model, requiring only 6 seconds per epoch with 0.46M parameters. However, its F1 score (82.95\%) remains within the mid-range of all evaluated methods.
(2) PatchTST, ResNet, and iTransformer are also parameter-efficient, but they achieve the lowest F1 scores, indicating limited modeling capacity for this task.
(3) TimesNet incurs the highest computational cost, requiring 57 seconds per epoch, yet its performance does not scale proportionally with its model size. 
(4) In general, multimodal approaches tend to be slower than unimodal models due to the additional cross-modal processing.
(5) In contrast, MedTVL achieves the highest F1 score, significantly outperforming all competing methods. 
The trainable parameters of MedTVL are 3.23M, which is larger than those of MedualTime, while remaining substantially more compact than LLM-based alternatives such as GPT4TS and MedTsLLM. Moreover, it runs faster than MedTsLLM, TimeVLM, and TimesNet.
(6) Overall, MedTVL strikes a favorable balance between effectiveness and efficiency, achieving superior performance with a manageable computational footprint.

\begin{table}[htb] % 双栏中的跨栏
%\vspace{-1mm} %表格和其上排版的距离
\caption{Backbone analysis in F1 score. 
%Full results are in \textbf{Appendix \ref{bac_app}}.
\textcolor{blue}{Blue}: second best in each module.
} %注意要求是在上方，还是表格下方
\label{tab:bac} %引用
\vspace{-4mm} % 表格内容和标题的距离
\centering % 正文居中
    
    \resizebox{\columnwidth}{!}{ % 盒子，表格占页面宽度
    %\scriptsize
    %\footnotesize
        \begin{threeparttable} %三栏式，可以在下面加标注
\begin{tabular}{>{\bfseries}l|l|c|c|c}
\midrule \midrule
\multicolumn{2}{c|}{\textbf{\makecell{Backbone of Module}}} & 
    \textbf{\makecell{APAVA\\(2-Classes)}} &
    \textbf{\makecell{TUSZ\\(4-Classes)}} &
    \textbf{\makecell{PTB-XL\\(4-Classes)}} \\
    
\midrule
\multirow{5}{*}{\rotatebox{90}{\makecell[c]{\small Temporal}}}
& ResNet (CNN) & 81.29\text{\scriptsize $\pm$1.56} & 86.09\text{\scriptsize $\pm$1.45} & {\color{blue}74.65\text{\scriptsize $\pm$1.67}} \\
%& ResCNN (CNN) & 80.99\text{\scriptsize $\pm$2.11} & {\color{blue}86.48\text{\scriptsize $\pm$1.11}} & 73.47\text{\scriptsize $\pm$2.13} \\
& TCN \cite{bai2018empirical} (CNN) &  {\color{blue}82.03\text{\scriptsize $\pm$0.89}} & 85.41\text{\scriptsize $\pm$1.98} & 74.37\text{\scriptsize $\pm$1.89} \\
& Medformer (Transformer) & 80.60\text{\scriptsize $\pm$1.98} & 84.66\text{\scriptsize $\pm$2.13} & 72.17\text{\scriptsize $\pm$2.34} \\
& PatchTST (Transformer) & 79.75\text{\scriptsize $\pm$2.45} & 85.19\text{\scriptsize $\pm$1.67} & 73.38\text{\scriptsize $\pm$1.56} \\
\midrule
% Visual通路（仅保留3个归属该通路的模型，multirow行数改为3）
\multirow{3}{*}{\rotatebox{90}{\makecell[c]{\small Visual}}}
& ViT \cite{ViT} (Transformer) & {\color{blue}82.59\text{\scriptsize $\pm$1.23}} & {\color{blue}84.17\text{\scriptsize $\pm$2.34}} & {\color{blue}75.17\text{\scriptsize $\pm$1.12}} \\
& ConvNeXts \cite{liu2022convnet} (CNN) & 79.71\text{\scriptsize $\pm$2.78} & 80.76\text{\scriptsize $\pm$2.56} & 69.65\text{\scriptsize $\pm$2.56} \\
& ResNet (CNN) & 78.53\text{\scriptsize $\pm$1.87} & 81.53\text{\scriptsize $\pm$1.89} & 71.64\text{\scriptsize $\pm$1.98} \\ \midrule
\multirow{2}{*}{\rotatebox{90}{\makecell[c]{\small Text}}}
& LLaMA (LLM) \cite{touvron2023llama} & {\color{blue}84.92\text{\scriptsize $\pm$2.15}} & {\color{blue}88.04\text{\scriptsize $\pm$1.78}}& {\color{blue}77.01\text{\scriptsize $\pm$1.67}} \\
& GPT-2 (LLM) \cite{radford2019gpt2} & 83.54\text{\scriptsize $\pm$1.87} & 87.55\text{\scriptsize $\pm$2.99} & 76.71\text{\scriptsize $\pm$2.45} \\ \midrule
\multicolumn{2}{c|}{\textbf{MedTVL}}  
& {\color{red} \textbf{85.52\text{\scriptsize $\pm$2.97}}} & {\color{red} \textbf{89.26\text{\scriptsize $\pm$0.82}}} & {\color{red} \textbf{77.93\text{\scriptsize $\pm$2.29}}} \\

% \multirow{5}{*}{\rotatebox{90}{\makecell[c]{Visual}}}
% & ViT (Transformer) & 82.59\text{\scriptsize $\pm$1.23} & 84.17\text{\scriptsize $\pm$2.34} & 75.17\text{\scriptsize $\pm$1.12} \\
% & Convnext (CNN) & 79.71\text{\scriptsize $\pm$2.78} & 80.76\text{\scriptsize $\pm$2.56} & 69.65\text{\scriptsize $\pm$2.56} \\
% & ResNet (CNN) & 78.53\text{\scriptsize $\pm$1.87} & 81.53\text{\scriptsize $\pm$1.89} & 71.64\text{\scriptsize $\pm$1.98} \\
% & MedTVL & {\color{red} \textbf{85.52\text{\scriptsize $\pm$0.79}}} & {\color{red} \textbf{89.26\text{\scriptsize $\pm$0.81}}} & {\color{red} \textbf{77.93\text{\scriptsize $\pm$0.92}}} \\
            %\bottomrule 
            \midrule \midrule
            \end{tabular} % 表尾
            
            %\begin{tablenotes} 
            %\item[1] Note: ...
            %\end{tablenotes}  % 注释，如果标题在下方，这个无法使用
        
        \end{threeparttable} % 三栏式，可以在下面加标注
                } % 盒子
\vspace{-3mm} %表格和其下排版的距离
\end{table} % 双栏中的跨栏
%\end{table} % 单栏
%\end{wraptable} %文字围绕表格
\textbf{(c) Backbone Analysis.}
In theory, the temporal, visual, and textual modules allow for multiple backbone instantiations. Nevertheless, in practice, the specific backbone choices play a critical role in shaping the framework’s overall effectiveness.
To examine the impact of backbone selection, we replace the default backbones with representative CNN, Transformer and LLM architectures. Table \ref{tab:bac} shows that different backbone choices lead to  different performance trends.
CNN-based backbones outperform Transformers in the temporal pathway, whereas the opposite holds true for the visual pathway. This phenomenon suggests that peak performance stems from a strategic heterogeneous architecture pairing. The inductive bias of CNNs is uniquely suited for capturing local, shift-invariant pathological motifs in temporal signals, while the Transformer's global modeling capability better aligns with the spatial structural correlations of 2D images. Within this optimized dual-stream framework, the selection of InceptionTime and Swin Transformer further enhances efficacy. Both models leverage multi-scale receptive fields to capture multi-granular medical signatures and exhibit superior computational efficiency—driven by Inception’s parallel convolutions and Swin’s shifted-window mechanism. This combination ensures a robust balance between diagnostic accuracy and inference latency. Additionally, the results of textual backbones show that general-purpose language models (e.g., GPT-2 and LLaMA) consistently degrades performance. This highlights the advantage of domain-specific clinical model (ClinicalBERT) in providing more effective semantic textual guidance.

%\begin{wrapfigure}{r}{0.45\textwidth}  % "r" 表示图像在右侧，0.5\textwidth 表示图像宽度为页面宽度的一半
% \begin{figure}[htb]
\begin{figure}[htb]
  %\vspace{-3mm}
  \begin{center}
    \includegraphics[width=0.45\textwidth]{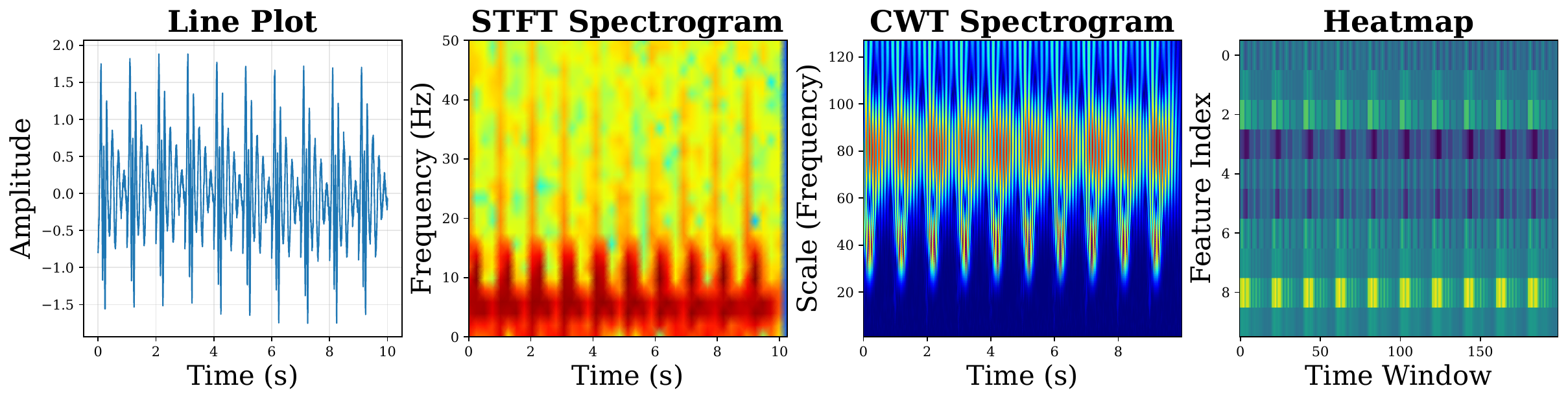} % 这里调整图像的宽度，确保它在wrapfigure中居中显示
  \end{center}
  \vspace{-3mm}
  \caption{An example of different visual representations.}
  %The bubble size represents the the model parameter size.
  % The dotted area denotes the amount of trainable parameters of different models.
  % Detection and classification tasks can be distinguished by marker type.}
    \label{fig:img}
    \vspace{-3mm} 
%\end{wrapfigure}
\end{figure}
\begin{figure}[h]
%\begin{figure*}[htb] % 双栏中的跨栏
%\begin{wrapfigure}{r}{0.6\textwidth}  % 'r' 表示右对齐，0.6\textwidth 表示图像将占据页面宽度的 60%。
%\vspace{-2mm}
\centering

    % 单个图片, 双栏中的单栏图width=0.49，双栏中的跨栏图 width=0.98
    %\includegraphics[width=0.49\textwidth, height=0.35\textheight]{./Figures/pdfs/AHGCSP.pdf}
    \includegraphics[width=0.45\textwidth]{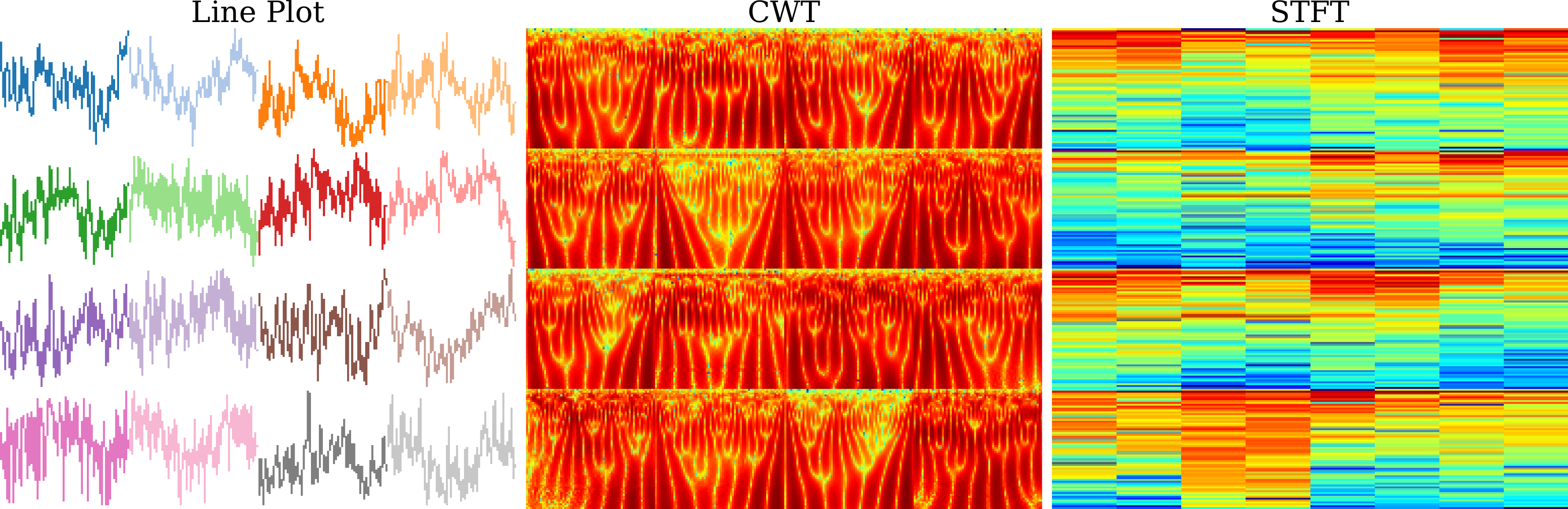}
      
    \vspace{-3mm} % 图片与标题的排版距离
    \caption{
    An example of grid construction of different visual representations on APAVA).}
    
    %  The red dashed line is MedSpaformer.
\label{fig:grid}
\vspace{-3mm} % 图片标题的下方排版距离
\end{figure}
%\end{figure*} % 双栏中的跨栏
%\end{wrapfigure}
\begin{table}[!ht] % 单栏排版
\caption{\textbf{Visual representation analysis}. %Full results see \textbf{Appendix \ref{img_app}}.
}
\label{tab:img} % 独立引用标签
\vspace{-4mm} % 表格与标题间距
\centering
\resizebox{\columnwidth}{!}{ % 适配单栏宽度
    \footnotesize
    \begin{threeparttable}
        \begin{tabular}{>{\bfseries}l|cc|cc|cc}
            \midrule \midrule
            % 表头：分两行合并，全加粗，与消融表结构一致
            \multicolumn{1}{l|}{\multirow{2}{*}{\begin{tabular}[c]{@{}l@{}}\textbf{Image}\\ \textbf{Representation}\end{tabular}}} &
            \multicolumn{2}{c|}{\textbf{APAVA (2-Classes)}} &
            \multicolumn{2}{c|}{\textbf{TUSZ (4-Classes)}} &
            \multicolumn{2}{c}{\textbf{PTB-XL (4-Classes)}} \\
            \multicolumn{1}{l|}{} &
            \textbf{F1} & \multicolumn{1}{c|}{\textbf{Acc.}} &
            \textbf{F1} & \multicolumn{1}{c|}{\textbf{Acc.}} &
            \textbf{F1} & \textbf{Acc.} \\ \midrule
            % 各图像表征：F1+Acc.带误差，标注最优/次优
            Line Plot  (grid) & \textcolor{red}{\textbf{87.11\text{\scriptsize $\pm$0.28}}} & \textcolor{red}{\textbf{87.77\text{\scriptsize $\pm$0.22}}} & 85.00\text{\scriptsize $\pm$0.32} & 89.64\text{\scriptsize $\pm$0.26} & 72.08\text{\scriptsize $\pm$0.25} & 80.72\text{\scriptsize $\pm$0.18} \\
            STFT   (grid)    & 82.74\text{\scriptsize $\pm$0.29} & 84.49\text{\scriptsize $\pm$0.33} & \textcolor{blue}{87.12\text{\scriptsize $\pm$0.20}} & \textcolor{blue}{89.98\text{\scriptsize $\pm$0.22}} & \textcolor{blue}{76.07\text{\scriptsize $\pm$0.34}} & \textcolor{blue}{83.73\text{\scriptsize $\pm$0.26}} \\
            Heatmap (grid)   & 83.02\text{\scriptsize $\pm$0.23} & 83.79\text{\scriptsize $\pm$0.26} & 86.28\text{\scriptsize $\pm$0.35} & 89.98\text{\scriptsize $\pm$0.29} & 75.01\text{\scriptsize $\pm$0.20} & 81.79\text{\scriptsize $\pm$0.31} \\ \midrule
            TimeVLM (non-grid) & 82.48\text{\scriptsize $\pm$1.56} & 83.02\text{\scriptsize $\pm$1.88} & 85.79\text{\scriptsize $\pm$0.89} & 90.50\text{\scriptsize $\pm$0.47} & 74.85\text{\scriptsize $\pm$1.92} & 81.46\text{\scriptsize $\pm$1.29} \\ \midrule
            % 主模型MedTVL：红体加粗标最优
            MedTVL (CWT)   & \textcolor{blue}{85.52\text{\scriptsize $\pm$2.97}} & 85.88\text{\scriptsize $\pm$1.85} & \textcolor{red}{\textbf{89.26\text{\scriptsize $\pm$0.82}}} & \textcolor{red}{\textbf{91.19\text{\scriptsize $\pm$2.50}}} & \textcolor{red}{\textbf{77.93\text{\scriptsize $\pm$2.29}}} & \textcolor{red}{\textbf{84.43\text{\scriptsize $\pm$0.79}}} \\
            \midrule \midrule
        \end{tabular}
    \end{threeparttable}
}
\vspace{-3mm} % 表格与下文间距
\end{table}

\textbf{(d) Visual Representation Analysis.}
We investigate the impact of different time-series-to-image representations on three datasets by replacing the CWT spectrogram in our grid structure with line plots, STFT spectrograms, and heatmaps (Figure \ref{fig:img} and \ref{fig:grid}). We further evaluate the role of grid construction by substituting the CWT-based grid with the Frequency–Periodicity–Multi-scale Convolution Encoding from TimeVLM \cite{Time-VLM}, which does not adopt a grid layout and is denoted as \textbf{“TimeVLM (non-grid)”}.
As shown in Table \ref{tab:img}, although CWT-based MedTVL does not achieve the best F1 score on every dataset, it delivers the strongest average performance across benchmarks, indicating greater robustness to diverse data characteristics. 
Line plots perform best on APAVA, likely because the relatively small number of time steps allows temporal details to be clearly represented in a line plot.
In contrast, the larger strides in TUSZ (6000) and PTB-XL (1000) lead to substantial loss of local patterns when using line plots. By explicitly modeling time–frequency structures and multi-scale dynamics, CWT remains effective across datasets. Moreover, grid-based representations consistently outperform the non-grid variant, highlighting the importance of structured temporal layouts.
%\begin{wrapfigure}{r}{0.45\textwidth}  % "r" 表示图像在右侧，0.5\textwidth 表示图像宽度为页面宽度的一半
% \begin{figure}[htb]
\begin{figure}[htb]
  %\vspace{-3mm}
  \begin{center}
    \includegraphics[width=0.45\textwidth]{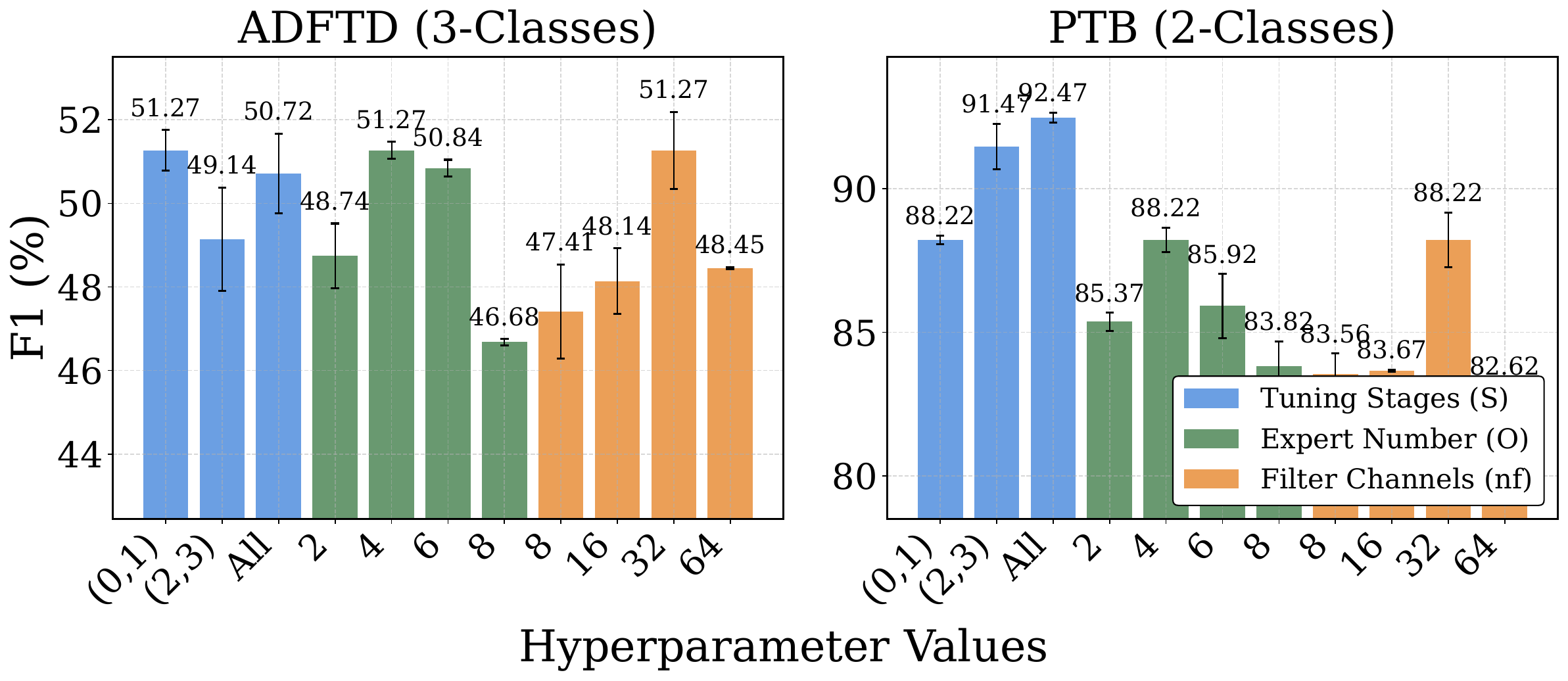} % 这里调整图像的宽度，确保它在wrapfigure中居中显示
  \end{center}
  \vspace{-3mm}
  \caption{Sensitivity Analysis in F1 of two datasets.}
  %The bubble size represents the the model parameter size.
  % The dotted area denotes the amount of trainable parameters of different models.
  % Detection and classification tasks can be distinguished by marker type.}
    \label{fig:hyper}
    \vspace{-3mm} 
%\end{wrapfigure}
\end{figure}

\textbf{(e) Sensitivity Analysis.}
We study the influence of three key hyperparameters on three datasets—ADFTD (3-Classes), TUSZ (4-Classes), and PTB (2-Classes): (1) tuning stages (S) of the visual backbone, with values ($[(0,1),(2,3), \text{all}]$); (2) filter channels $nf$ of the temporal backbone, with values ($[8,16,32,64]$); and (3) expert number $O$ in the MoE module, with values ($[2,4,6,8]$). 
%Full results are in \textbf{Appendix \ref{hyper_app}}, and 
Figure \ref{fig:hyper} presents F1 on two datasets.
For tuning stages, ADFTD achieves its best performance when only the first two stages are fine-tuned, whereas PTB reaches its optimum when all stages are tuned. Since full tuning incurs higher computational cost, tuning only the first two stages provides a favorable trade-off between performance and efficiency.
Regarding the expert number $O$, both datasets attain peak performance at ($O=4$). Increasing $O$ further leads to performance degradation, likely because too many experts dilute the training data per expert and introduce redundancy, harming generalization.
For the filter channel number $nf$, a value of 32 consistently yields the best results across datasets, indicating an appropriate balance between temporal modeling capacity and model complexity. Smaller values limit expressiveness, while larger ones tend to introduce redundancy without clear performance gains.

%\begin{wrapfigure}{r}{0.45\textwidth}  % "r" 表示图像在右侧，0.5\textwidth 表示图像宽度为页面宽度的一半
% \begin{figure}[htb]
\begin{figure}[htb]
  \vspace{-3mm}
  \begin{center}
    \includegraphics[width=0.45\textwidth]{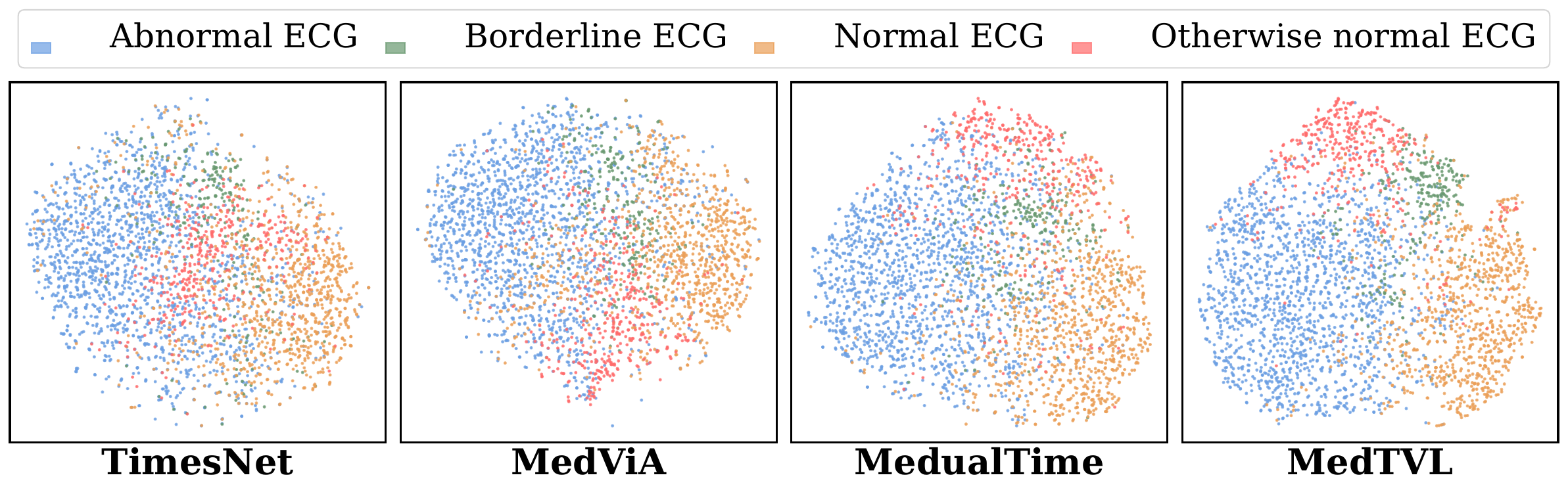} % 这里调整图像的宽度，确保它在wrapfigure中居中显示
  \end{center}
  \vspace{-3mm}
  \caption{Visualization analysis of PTB-XL(4-Classes).}
  %The bubble size represents the the model parameter size.
  % The dotted area denotes the amount of trainable parameters of different models.
  % Detection and classification tasks can be distinguished by marker type.}
    \label{fig:visual}
    \vspace{-3mm} 
%\end{wrapfigure}
\end{figure}
\textbf{(f) Visualization.}
To provide an intuitive visualization of the representations learned under supervised training, we apply t-SNE \cite{maaten2008visualizing} to project the embeddings of representative models on the PTB-XL (4-Classes) dataset into a two-dimensional space, as shown in Figure \ref{fig:visual}.
The results illustrate a progressive improvement in representation separability across models. TimesNet primarily distinguishes the two largest classes, while MedViA and MedualTime further enhance class separability, particularly for the third major class. In contrast, MedTVL produces the most structured embedding space, with all classes forming well-separated clusters.

\section{Conclusion}
This work presents MedTVL, a tri-modal framework for medical time series classification that integrates numerical signals, visual representations, and clinical language to better reflect clinical practice. By adopting a heterogeneous dual-pathway design, MedTVL leverages complementary inductive biases from convolutional and transformer-based architectures to capture both fine-grained temporal dynamics in numerical signals and global morphological patterns in time-series-derived images. Furthermore, the dual-pathway structure naturally enables cross-modal positive pairing, facilitating multimodal contrastive learning under limited clinical annotations. Extensive experiments across diverse tasks demonstrate the robustness, adaptability, and generalizability of MedTVL for clinical decision support. 
%The limitations and future directions are discussed in Appendix \ref{lim_app}.

\clearpage
\pagebreak

\section{Acknowledgements}
This work is funded by National Natural Science Foundation of China Grant No. 72371217,  NSFC 62572418 and Guangdong Provincial Talent Program (No.2024TQ08X366), the Guangzhou Industrial Informatics and Intelligence Key Laboratory No. 2024A03J0628, the Nansha Key Area Science and Technology Project No. 2023ZD003, and Project No. 2021JC02X191.

%This work is supported by NSFC 62572418 and Guangdong Provincial Talent Program(No.2024TQ08X366）

\section*{Limitations and Ethical Considerations}
This study uses publicly available, de-identified medical datasets from prior research and does not involve the collection of new data from human participants. As the data are fully anonymized, no additional Institutional Review Board (IRB) approval is required under our institutional guidelines. All experiments comply with ACM Publications Policies, including those on research involving human participants.
MedTVL is intended solely as a research and decision-support framework rather than a standalone diagnostic system. Its predictions should be interpreted by qualified professionals and must not replace clinical judgment, as misuse without appropriate domain expertise may lead to incorrect decisions.
Despite strong performance and transferability across multiple datasets and learning paradigms, MedTVL remains subject to limitations stemming from dataset quality, representativeness, and potential biases. While developed for medical applications, its general design principles may extend to other domains; any such use should undergo domain-specific ethical review and regulatory oversight. We acknowledge that additional ethical concerns may be raised by the community.

% This work advances multimodal learning for medical data analysis. The publicly available, de-identified dataset from prior research ensures compliance with privacy regulations and ethical standards, with no patient privacy concerns. 
% By integrating time series, visual, and textual modalities within a unified framework, the proposed approach advances multimodal learning for medical decision support while remaining broadly applicable beyond healthcare. Non-professionals should use it as auxiliary tool. 
% The design principles of MedTVL can be readily extended to other domains where numerical signals, visual patterns, and contextual knowledge jointly inform decision-making, such as smart manufacturing, and human–machine interaction. Moreover, the incorporation of multimodal contrastive learning alleviates reliance on large-scale expert annotations, lowering the barrier to deploying advanced models in data-scarce settings. Overall, this work contributes a generalizable and ethically responsible multimodal paradigm that promotes more robust learning across real-world applications.

%%
%% The next two lines define the bibliography style to be used, and
%% the bibliography file.
\bibliographystyle{ACM-Reference-Format}
\balance
\bibliography{ref}

@article{wu2024srt,
  title={SRT: Improved transformer-based model for classification of 2D heartbeat images},
  author={Wu, Wenwen and Huang, Yanqi and Wu, Xiaomei},
  journal={Biomedical Signal Processing and Control},
  volume={88},
  pages={105017},
  year={2024},
  publisher={Elsevier}
}

@inproceedings{wang2017time,
  title={Time series classification from scratch with deep neural networks: A strong baseline},
  author={Wang, Zhiguang and Yan, Weizhong and Oates, Tim},
  booktitle={2017 International joint conference on neural networks (IJCNN)},
  pages={1578--1585},
  year={2017}
}

@inproceedings{he2016deep,
  title={Deep residual learning for image recognition},
  author={He, Kaiming and Zhang, Xiangyu and Ren, Shaoqing and Sun, Jian},
  booktitle={Proceedings of the IEEE conference on computer vision and pattern recognition},
  pages={770--778},
  year={2016}
}

@article{zhao2017convolutional,
  title={Convolutional neural networks for time series classification},
  author={Zhao, Bendong and Lu, Huanzhang and Chen, Shangfeng and Liu, Junliang and Wu, Dongya},
  journal={Journal of systems engineering and electronics},
  volume={28},
  number={1},
  pages={162--169},
  year={2017}
}

@article{InceptionTime,
  title={Inceptiontime: Finding alexnet for time series classification},
  author={Ismail Fawaz, Hassan and Lucas, Benjamin and Forestier, Germain and Pelletier, Charlotte and Schmidt, Daniel F and Weber, Jonathan and Webb, Geoffrey I and Idoumghar, Lhassane and Muller, Pierre-Alain and Petitjean, Fran{\c{c}}ois},
  journal={Data Mining and Knowledge Discovery},
  volume={34},
  number={6},
  pages={1936--1962},
  year={2020}
}

@article{cui2016multi,
  title={Multi-scale convolutional neural networks for time series classification},
  author={Cui, Zhicheng and Chen, Wenlin and Chen, Yixin},
  journal={arXiv preprint arXiv:1603.06995},
  year={2016}
}

@article{almanza2023emotion,
  title={Emotion recognition in EEG signals using the continuous wavelet transform and CNNs},
  author={Almanza-Conejo, Oscar and Almanza-Ojeda, Dora Luz and Contreras-Hernandez, Jose Luis and Ibarra-Manzano, Mario Alberto},
  journal={Neural Computing and Applications},
  volume={35},
  number={2},
  pages={1409--1422},
  year={2023}
}

@article{yun2025temporal,
  title={Temporal dynamics and biological variability of Alzheimer biomarkers},
  author={Yun, Jihwan and Shin, Daeun and Lee, Eun Hye and Kim, Jun Pyo and Ham, Hongki and Gu, Yuna and Chun, Min Young and Kang, Sung Hoon and Kim, Hee Jin and Na, Duk L and others},
  journal={JAMA neurology},
  volume={82},
  number={4},
  pages={384--396},
  year={2025}
}

@article{mu2025comprehensive,
  title={A comprehensive survey of mixture-of-experts: Algorithms, theory, and applications},
  author={Mu, Siyuan and Lin, Sen},
  journal={arXiv preprint arXiv:2503.07137},
  year={2025}
}

@article{liu2025mofe,
  title={Mofe-time: mixture of frequency domain experts for time-series forecasting models},
  author={Liu, Yiwen and Zhang, Chenyu and Song, Junjie and Chen, Siqi and Yin, Sun and Wang, Zihan and Zeng, Lingming and Cao, Yuji and Jiao, Junming},
  journal={arXiv preprint arXiv:2507.06502},
  year={2025}
}

@inproceedings{liumoirai,
  title={Moirai-MoE: Empowering Time Series Foundation Models with Sparse Mixture of Experts},
  author={Liu, Xu and Liu, Juncheng and Woo, Gerald and Aksu, Taha and Liang, Yuxuan and Zimmermann, Roger and Liu, Chenghao and Li, Junnan and Savarese, Silvio and Xiong, Caiming and others},
  booktitle={Forty-second International Conference on Machine Learning},
  year={2025}
}

@inproceedings{Time-MoE,
  title={Time-MoE: Billion-Scale Time Series Foundation Models with Mixture of Experts},
  author={Shi, Xiaoming and Wang, Shiyu and Nie, Yuqi and Li, Dianqi and Ye, Zhou and Wen, Qingsong and Jin, Ming},
  booktitle={The Thirteenth International Conference on Learning Representations},
  year={2025}
}

@article{mclaren2024st,
  title={From ST-segment elevation MI to occlusion MI: the new paradigm shift in acute myocardial infarction},
  author={McLaren, Jesse and de Alencar, Jos{\'e} Nunes and Aslanger, Emre K and Meyers, H Pendell and Smith, Stephen W},
  journal={JACC: Advances},
  volume={3},
  number={11},
  pages={101314},
  year={2024}
}

@article{klewer2022premature,
  title={Premature ventricular contractions (PVCs): a narrative review},
  author={Klewer, Jake and Springer, Jennifer and Morshedzadeh, Jack},
  journal={The American Journal of Medicine},
  volume={135},
  number={11},
  pages={1300--1305},
  year={2022}
}

@article{woo2024unified,
  title={Unified Training of Universal Time Series Forecasting Transformers},
  author={Woo, Gerald and Liu, Chenghao and Kumar, Akshat and Xiong, Caiming and Savarese, Silvio and Sahoo, Doyen},
  journal={Proceedings of Machine Learning Research},
  volume={235},
  pages={53140--53164},
  year={2024}
}

@article{gu2025foundation,
  title={Foundation Models for Biosignals: A Survey},
  author={Gu, Xiao and Shu, Yuxuan and Han, Jinpei and Liu, Yuxuan and Liu, Zhangdaihong and Anibal, James and Sangha, Veer and Phillips, Edward and Segal, Bradley and Yuan, Hang and others},
  journal={Authorea Preprints},
  year={2025}
}

@article{ding2025advances,
  title={Advances in deep learning for personalized ECG diagnostics: A systematic review addressing inter-patient variability and generalization constraints},
  author={Ding, Cheng and Yao, Tianliang and Wu, Chenwei and Ni, Jianyuan},
  journal={Biosensors and Bioelectronics},
  volume={271},
  pages={117073},
  year={2025},
  publisher={Elsevier}
}

@article{sharma2024emerging,
  title={Emerging trends in EEG signal processing: A systematic review},
  author={Sharma, Ramnivas and Meena, Hemant Kumar},
  journal={SN Computer Science},
  volume={5},
  number={4},
  pages={415},
  year={2024},
  publisher={Springer}
}

@article{rim2020deep,
  title={Deep learning in physiological signal data: A survey},
  author={Rim, Beanbonyka and Sung, Nak-Jun and Min, Sedong and Hong, Min},
  journal={Sensors},
  volume={20},
  number={4},
  pages={969},
  year={2020}
}

@article{fatourechi2007emg,
  title={EMG and EOG artifacts in brain computer interface systems: A survey},
  author={Fatourechi, Mehrdad and Bashashati, Ali and Ward, Rabab K and Birch, Gary E},
  journal={Clinical neurophysiology},
  volume={118},
  number={3},
  pages={480--494},
  year={2007}
}

@inproceedings{LiuHZWWML24,
  author       = {Yong Liu and
                  Tengge Hu and
                  Haoran Zhang and
                  Haixu Wu and
                  Shiyu Wang and
                  Lintao Ma and
                  Mingsheng Long},
  title        = {iTransformer: Inverted Transformers Are Effective for Time Series
                  Forecasting},
  booktitle    = {International Conference on Learning Representations},
  year         = {2024}
}

@inproceedings{Tang2021SelfSupervisedGN,
  title={Self-Supervised Graph Neural Networks for Improved Electroencephalographic Seizure Analysis},
  author={Siyi Tang and Jared A. Dunnmon and Khaled Saab and Xuan Zhang and Qianying Huang and Florian Dubost and D. Rubin and Christopher Lee-Messer},
  booktitle={International Conference on Learning Representations},
  year={2021}
}

@article{escudero2006analysis,
  title={Analysis of electroencephalograms in Alzheimer's disease patients with multiscale entropy},
  author={Escudero, J and Ab{\'a}solo, Daniel and Hornero, Roberto and Espino, Pedro and L{\'o}pez, Miguel},
  journal={Physiological measurement},
  volume={27},
  number={11},
  pages={1091},
  year={2006}
}

@article{miltiadous2023dataset,
  title={A dataset of scalp EEG recordings of Alzheimer’s disease, frontotemporal dementia and healthy subjects from routine EEG},
  author={Miltiadous, Andreas and Tzimourta, Katerina D and Afrantou, Theodora and Ioannidis, Panagiotis and Grigoriadis, Nikolaos and Tsalikakis, Dimitrios G and Angelidis, Pantelis and Tsipouras, Markos G and Glavas, Euripidis and Giannakeas, Nikolaos and others},
  journal={Data},
  volume={8},
  number={6},
  pages={95},
  year={2023}
}

@article{physiobank2000physionet,
  title={Physionet: components of a new research resource for complex physiologic signals},
  author={PhysioBank, PhysioToolkit},
  journal={Circulation},
  volume={101},
  number={23},
  pages={e215--e220},
  year={2000}
}

@article{wagner2020ptb,
  title={PTB-XL, a large publicly available electrocardiography dataset},
  author={Wagner, Patrick and Strodthoff, Nils and Bousseljot, Ralf-Dieter and Kreiseler, Dieter and Lunze, Fatima I and Samek, Wojciech and Schaeffter, Tobias},
  journal={Scientific data},
  volume={7},
  number={1},
  pages={1--15},
  year={2020}
}

@article{shah2018temple,
  title={The temple university hospital seizure detection corpus},
  author={Shah, Vinit and Von Weltin, Eva and Lopez, Silvia and McHugh, James Riley and Veloso, Lillian and Golmohammadi, Meysam and Obeid, Iyad and Picone, Joseph},
  journal={Frontiers in neuroinformatics},
  volume={12},
  pages={83},
  year={2018}
}

@inproceedings{zeng2023dlinear,
  title={Are transformers effective for time series forecasting?},
  author={Zeng, Ailing and Chen, Muxi and Zhang, Lei and Xu, Qiang},
  booktitle={Proceedings of the AAAI conference on artificial intelligence},
  volume={37},
  
  pages={11121--11128},
  year={2023}
}

@inproceedings{nie2023time,
  title={A Time Series is Worth 64 Words: Long-term Forecasting with Transformers},
  author={Nie, Yuqi and Nguyen, Nam H. and Sinthong, Phanwadee and Kalagnanam, Jayant},
  booktitle={International Conference on Learning Representations},
  year={2023}
}

@inproceedings{wu2022timesnet,
  title={Timesnet: Temporal 2d-variation modeling for general time series analysis},
  author={Wu, Haixu and Hu, Tengge and Liu, Yong and Zhou, Hang and Wang, Jianmin and Long, Mingsheng},
  booktitle={International Conference on Learning Representations},
  year={2022}
}

@article{tan2022multirocket,
  title={MultiRocket: multiple pooling operators and transformations for fast and effective time series classification},
  author={Tan, Chang Wei and Dempster, Angus and Bergmeir, Christoph and Webb, Geoffrey I},
  journal={Data Mining and Knowledge Discovery},
  volume={36},
  number={5},
  pages={1623--1646},
  year={2022}
}

@inproceedings{Salloum_Kuo_2017,   title={ECG-based biometrics using recurrent neural networks},  booktitle={ICASSP},  author={Salloum, Ronald and Kuo, C.-C. Jay},  year={2017},  month={Mar},  pages={2062–2066}}

@article{Lawhern_2018,   title={EEGNet: A Compact Convolutional Network for EEG-based Brain-Computer Interfaces},  journal={Journal of Neural Engineering},  author={Lawhern, Vernon J and Solon, Amelia J and Waytowich, Nicholas R and Gordon, Stephen M and Hung, Chou P and Lance, Brent J},  year={2018},  month={Oct},  pages={056013} }

@article{Medformer,
  title={Medformer: A multi-granularity patching transformer for medical time-series classification},
  author={Wang, Yihe and Huang, Nan and Li, Taida and Yan, Yujun and Zhang, Xiang},
  journal={Advances in Neural Information Processing Systems},
  volume={37},
  pages={36314--36341},
  year={2024}
}

@article{ye2024empowering,
  title={Empowering Time Series Analysis with Foundation Models: A Comprehensive Survey},
  author={Ye, Jiexia and Yu, Yongzi and Zhang, Weiqi and Wang, Le and Li, Jia and Tsung, Fugee},
  journal={arXiv preprint arXiv:2405.02358},
  year={2024}
}

@inproceedings{MedSpaformer,
  title={MedSpaformer: a Transferable Transformer with Multi-granularity Token Sparsification for Medical Time Series Classification},
  author={Ye, Jiexia and Zhang, Weiqi and Li, Ziyue and Li, Jia and Tsung, Fugee},
  booktitle={Proceedings of the AAAI Conference on Artificial Intelligence},
  volume={40},
  number={33},
  pages={27791--27799},
  year={2026}
}

@inproceedings{pratiher2022dilated,
  title={A dilated residual vision transformer for atrial fibrillation detection from stacked time-frequency ecg representations},
  author={Pratiher, Sawon and Srivastava, Apoorva and Priyatha, Yedla Bindu and Ghosh, Nirmalya and Patra, Amit},
  booktitle={ICASSP 2022-2022 IEEE International Conference on Acoustics, Speech and Signal Processing (ICASSP)},
  pages={1121--1125},
  year={2022},
  organization={IEEE}
}

@article{GPT4TS,
  title={One fits all: Power general time series analysis by pretrained lm},
  author={Zhou, Tian and Niu, Peisong and Sun, Liang and Jin, Rong and others},
  journal={Advances in neural information processing systems},
  volume={36},
  year={2024}
}

@inproceedings{jin2023timellm,
  title={Time-LLM: Time Series Forecasting by Reprogramming Large Language Models},
  author={Jin, Ming and Wang, Shiyu and Ma, Lintao and Chu, Zhixuan and Zhang, James Y and Shi, Xiaoming and Chen, Pin-Yu and Liang, Yuxuan and Li, Yuan-Fang and Pan, Shirui and others},
  booktitle={The Twelfth International Conference on Learning Representations},
  year={2023}
}

@article{gao2024units,
  title={Units: A unified multi-task time series model},
  author={Gao, Shanghua and Koker, Teddy and Queen, Owen and Hartvigsen, Tom and Tsiligkaridis, Theodoros and Zitnik, Marinka},
  journal={Advances in Neural Information Processing Systems},
  volume={37},
  pages={140589--140631},
  year={2024}
}

@article{cheng2024advancing,
  title={Advancing Time Series Classification with Multimodal Language Modeling},
  author={Cheng, Mingyue and Chen, Yiheng and Liu, Qi and Liu, Zhiding and Luo, Yucong},
  journal={arXiv preprint arXiv:2403.12371},
  year={2024}
}

@inproceedings{MedGNN,
  title={Towards multi-resolution spatiotemporal graph learning for medical time series classification},
  author={Fan, Wei and Fei, Jingru and Guo, Dingyu and Yi, Kun and Song, Xiaozhuang and Xiang, Haolong and Ye, Hangting and Li, Min},
  booktitle={Proceedings of the ACM on Web Conference 2025},
  pages={5054--5064},
  year={2025}
}

@article{MedTsLLM,
  title={MedTsLLM: Leveraging LLMs for Multimodal Medical Time Series Analysis},
  author={Chan, Nimeesha and Parker, Felix and Bennett, William and Wu, Tianyi and Jia, Mung Yao and Fackler, James and Ghobadi, Kimia},
  journal={arXiv preprint arXiv:2408.07773},
  year={2024}
}

@inproceedings{MedualTime,
  author       = {Jiexia Ye and
                  Weiqi Zhang and
                  Ziyue Li and
                  Jia Li and
                  Meng Zhao and
                  Fugee Tsung},
  title        = {MedualTime: {A} Dual-Adapter Language Model for Medical Time Series-Text
                  Multimodal Learning},
  booktitle    = {Proceedings of the Thirty-Fourth International Joint Conference on
                  Artificial Intelligence, {IJCAI} 2025, Montreal, Canada, August 16-22,
                  2025},
  pages        = {7913--7921},
  publisher    = {ijcai.org},
  year         = {2025}
}

@article{liu2023self,
  title={Self-supervised contrastive learning for medical time series: A systematic review},
  author={Liu, Ziyu and Alavi, Azadeh and Li, Minyi and Zhang, Xiang},
  journal={Sensors},
  volume={23},
  number={9},
  pages={4221},
  year={2023}
}

@inproceedings{MERL,
  title={Zero-Shot ECG Classification with Multimodal Learning and Test-time Clinical Knowledge Enhancement},
  author={Liu, Che and Wan, Zhongwei and Ouyang, Cheng and Shah, Anand and Bai, Wenjia and Arcucci, Rossella},
  booktitle={Forty-first International Conference on Machine Learning},
  year = {2024}
}

@inproceedings{METS,
  title={Frozen language model helps ECG zero-shot learning},
  author={Li, Jun and Liu, Che and Cheng, Sibo and Arcucci, Rossella and Hong, Shenda},
  booktitle={Medical Imaging with Deep Learning},
  pages={402--415},
  year={2024}
}

@inproceedings{AimTS,
  title={AimTS: Augmented Series and Image Contrastive Learning for Time Series Classification},
  author={Chen, Yuxuan and Huang, Shanshan and Cheng, Yunyao and Chen, Peng and Rao, Zhongwen and Shu, Yang and Yang, Bin and Pan, Lujia and Guo, Chenjuan},
  booktitle={2025 IEEE 41st International Conference on Data Engineering (ICDE)},
  pages={1952--1965},
  year={2025}
}

@inproceedings{TimesURL,
  title={Timesurl: Self-supervised contrastive learning for universal time series representation learning},
  author={Liu, Jiexi and Chen, Songcan},
  booktitle={Proceedings of the AAAI conference on artificial intelligence},
  volume={38},
  number={12},
  pages={13918--13926},
  year={2024}
}

@article{T-Loss,
  title={Unsupervised scalable representation learning for multivariate time series},
  author={Franceschi, Jean-Yves and Dieuleveut, Aymeric and Jaggi, Martin},
  journal={Advances in neural information processing systems},
  volume={32},
  year={2019}
}

@inproceedings{TS-TCC,
  title={Time-Series Representation Learning via Temporal and Contextual Contrasting},
  author={Eldele, Emadeldeen and Ragab, Mohamed and Chen, Zhenghua and Wu, Min and Kwoh, Chee Keong and Li, Xiaoli and Guan, Cuntai},
  booktitle={Proceedings of the Thirtieth International Joint Conference on Artificial Intelligence},
  pages={2352--2359},
  year={2021}
}

@inproceedings{InfoTS,
  title={Time series contrastive learning with information-aware augmentations},
  author={Luo, Dongsheng and Cheng, Wei and Wang, Yingheng and Xu, Dongkuan and Ni, Jingchao and Yu, Wenchao and Zhang, Xuchao and Liu, Yanchi and Chen, Yuncong and Chen, Haifeng and others},
  booktitle={Proceedings of the Thirty-Seventh AAAI Conference on Artificial Intelligence and Thirty-Fifth Conference on Innovative Applications of Artificial Intelligence and Thirteenth Symposium on Educational Advances in Artificial Intelligence},
  pages={4534--4542},
  year={2023}
}

@inproceedings{Ts2vec,
  title={Ts2vec: Towards universal representation of time series},
  author={Yue, Zhihan and Wang, Yujing and Duan, Juanyong and Yang, Tianmeng and Huang, Congrui and Tong, Yunhai and Xu, Bixiong},
  booktitle={Proceedings of the AAAI conference on artificial intelligence},
  volume={36},
  number={8},
  pages={8980--8987},
  year={2022}
}

@inproceedings{tonekaboniunsupervised,
  title={Unsupervised Representation Learning for Time Series with Temporal Neighborhood Coding},
  author={Tonekaboni, Sana and Eytan, Danny and Goldenberg, Anna},
  booktitle={International Conference on Learning Representations},
  year = {2021}
}

@inproceedings{woocost,
  title={CoST: Contrastive Learning of Disentangled Seasonal-Trend Representations for Time Series Forecasting},
  author={Woo, Gerald and Liu, Chenghao and Sahoo, Doyen and Kumar, Akshat and Hoi, Steven},
  booktitle={International Conference on Learning Representations},
  year = {2021}
}

@article{zhang2022self,
  title={Self-supervised contrastive pre-training for time series via time-frequency consistency},
  author={Zhang, Xiang and Zhao, Ziyuan and Tsiligkaridis, Theodoros and Zitnik, Marinka},
  journal={Advances in neural information processing systems},
  volume={35},
  pages={3988--4003},
  year={2022}
}

@inproceedings{huang2016deep,
  title={Deep networks with stochastic depth},
  author={Huang, Gao and Sun, Yu and Liu, Zhuang and Sedra, Daniel and Weinberger, Kilian Q},
  booktitle={European conference on computer vision},
  pages={646--661},
  year={2016},
  organization={Springer}
}

@article{han2021transformer,
  title={Transformer in transformer},
  author={Han, Kai and Xiao, An and Wu, Enhua and Guo, Jianyuan and Xu, Chunjing and Wang, Yunhe},
  journal={Advances in neural information processing systems},
  volume={34},
  pages={15908--15919},
  year={2021}
}

@inproceedings{liu2021swin,
  title={Swin transformer: Hierarchical vision transformer using shifted windows},
  author={Liu, Ze and Lin, Yutong and Cao, Yue and Hu, Han and Wei, Yixuan and Zhang, Zheng and Lin, Stephen and Guo, Baining},
  booktitle={Proceedings of the IEEE/CVF international conference on computer vision},
  pages={10012--10022},
  year={2021}
}

@article{bai2018empirical,
  title={An Empirical Evaluation of Generic Convolutional and Recurrent Networks for Sequence Modeling},
  author={Bai, Shaojie},
  journal={arXiv preprint arXiv:1803.01271},
  year={2018}
}

@inproceedings{liu2022convnet,
  title={A convnet for the 2020s},
  author={Liu, Zhuang and Mao, Hanzi and Wu, Chao-Yuan and Feichtenhofer, Christoph and Darrell, Trevor and Xie, Saining},
  booktitle={Proceedings of the IEEE/CVF conference on computer vision and pattern recognition},
  pages={11976--11986},
  year={2022}
}

@inproceedings{ViT,
  title={An Image is Worth 16x16 Words: Transformers for Image Recognition at Scale},
  author={Dosovitskiy, Alexey and Beyer, Lucas and Kolesnikov, Alexander and Weissenborn, Dirk and Zhai, Xiaohua and Unterthiner, Thomas and Dehghani, Mostafa and Minderer, Matthias and Heigold, Georg and Gelly, Sylvain and others},
  booktitle={International Conference on Learning Representations},
  year={2020}
}

@inproceedings{CLIP,
  title     = {Learning Transferable Visual Models From Natural Language Supervision},
  author    = {Radford, Alec and Kim, Jong Wook and Hallacy, Chris and Ramesh, Aditya and Goh, Gabriel and Agarwal, Sandhini and Sastry, Girish and Askell, Amanda and Mishkin, Pamela and Clark, Jack and Krueger, Gretchen and Sutskever, Ilya},
  booktitle = {Proceedings of the 38th International Conference on Machine Learning (ICML)},
  series    = {Proceedings of Machine Learning Research},
  volume    = {139},
  pages     = {8748--8763},
  year      = {2021}
}

@article{OccamVTS,
  title={OccamVTS: Distilling Vision Models to 1\% Parameters for Time Series Forecasting},
  author={Lyu, Sisuo and Zhong, Siru and Ruan, Weilin and Liu, Qingxiang and Wen, Qingsong and Xiong, Hui and Liang, Yuxuan},
  journal={arXiv preprint arXiv:2508.01727},
  year={2025}
}

@inproceedings{VisionTS,
  author    = {Mouxiang Chen and Lefei Shen and Zhuo Li and Xiaoyun Joy Wang and Jianling Sun and Chenghao Liu},
  title     = {VisionTS: Visual Masked Autoencoders Are Free-Lunch Zero-Shot Time Series Forecasters},
  booktitle = {Proceedings of the 42nd International Conference on Machine Learning (ICML)},
  year      = {2025}
}

@inproceedings{ViTST,
  author  = {Zeyu Li and Shaopeng Li and Xueqian Yan},
  title   = {Time Series as Images: Vision Transformer for Irregularly Sampled Time Series},
  booktitle = {Advances in Neural Information Processing Systems},
  year    = {2023},
  volume  = {36},
  pages   = {49187--49204}
}

@article{DMMV,
  title={Multi-Modal View Enhanced Large Vision Models for Long-Term Time Series Forecasting},
  author={Shen, ChengAo and Yu, Wenchao and Zhao, Ziming and Song, Dongjin and Cheng, Wei and Chen, Haifeng and Ni, Jingchao},
  journal={arXiv preprint arXiv:2505.24003},
  year={2025}
}

@inproceedings{Time-VLM,
  title={Time-VLM: Exploring Multimodal Vision-Language Models for Augmented Time Series Forecasting},
  author={Zhong, Siru and Ruan, Weilin and Jin, Ming and Li, Huan and Wen, Qingsong and Liang, Yuxuan},
  booktitle={International Conference on Machine Learning (ICML), Poster},
  year={2025}
}

@article{GEM,
  title={Gem: Empowering mllm for grounded ecg understanding with time series and images},
  author={Lan, Xiang and Wu, Feng and He, Kai and Zhao, Qinghao and Hong, Shenda and Feng, Mengling},
  journal={arXiv preprint arXiv:2503.06073},
  year={2025}
}

@article{MedViA,
  title={MedViA: Empowering medical time series classification with vision augmentation and multimodal fusion},
  author={Fan, Wei and Fei, Jingru and Han, Jindong and Lian, Jie and Ye, Hangting and Song, Xiaozhuang and Lv, Xin and Yi, Kun and Li, Min},
  journal={Information Fusion},
  pages={103659},
  year={2025}
}

@article{touvron2023llama,
  title={Llama: Open and efficient foundation language models},
  author={Touvron, Hugo and Lavril, et.al},
  journal={arXiv preprint arXiv:2302.13971},
  year={2023}
}

@article{radford2019gpt2,
  title={Language models are unsupervised multitask learners},
  author={Radford, Alec and Wu, Jeffrey and Child, Rewon and Luan, David and Amodei, Dario and Sutskever, Ilya and others},
  journal={OpenAI blog},
  volume={1},
  number={8},
  pages={9},
  year={2019}
}

@article{wang2023optimized,
  title={Optimized glycemic control of type 2 diabetes with reinforcement learning: a proof-of-concept trial},
  author={Wang, Guangyu and Liu, Xiaohong and Ying, Zhen and Yang, Guoxing and Chen, Zhiwei and Liu, Zhiwen and Zhang, Min and Yan, Hongmei and Lu, Yuxing and Gao, Yuanxu and others},
  journal={Nature Medicine},
  volume={29},
  number={10},
  pages={2633--2642},
  year={2023}
}

@article{maaten2008visualizing,
  title={Visualizing data using t-SNE},
  author={Maaten, Laurens van der and Hinton, Geoffrey E},
  journal={Journal of Machine Learning Research},
  volume={9},
  pages={2579--2605},
  year={2008}
}

@inproceedings{dixit2024vision,
  title={Vision Language Models Are Few-Shot Audio Spectrogram Classifiers},
  author={Dixit, Satvik and Heller, Laurie and Donahue, Chris},
  booktitle={Audio Imagination: NeurIPS 2024 Workshop AI-Driven Speech, Music, and Sound Generation}
}
%\newpage
%\appendix

\section{Appendix}

% 只保留F1
%\begin{wraptable}{r}{0.6\textwidth} %文字围绕表格
%\begin{table}[htb] % 单栏
\begin{table*}[htb] % 双栏中的跨栏
%\vspace{-4mm} %表格和其上排版的距离
\caption{ Summary of benchmark datasets, including 
dataset statistics, train-validation-test splits, and data urls.} %注意要求是在上方，还是表格下方
\label{tab:data_app}
\centering % 正文居中
%\vspace{-3mm} % 表格内容和标题的距离
    \resizebox{\textwidth}{!}{ % 盒子，表格占页面宽度
    %\scriptsize
    \footnotesize
        \begin{threeparttable} %三栏式，可以在下面加标注
\begin{tabular}{l|l|r|c|r|r|r|r|r}
\midrule \midrule
\textbf{Datasets} & \textbf{Diseases} & \textbf{Total Samples} & \textbf{Classes} & \textbf{Channels} & \textbf{Steps} & \textbf{Training} & \textbf{Validation} & \textbf{Test} \\ 
\midrule
APAVA (2-Classes)   & Alzheimer & 5,967  & 2 & 16 & 256  & 3,123  & 1,413  & 1,431  \\ 
\midrule
ADFTD (3-Classes)   & Alzheimer & 69,752 & 3 & 19 & 256  & 40,446 & 14,658 & 14,648 \\ 
\midrule
TUSZ (2-Classes)    & Epilepsy & 22,040 & 2 & 19 & 6,000 & 13,224 & 4,408  & 4,408  \\ 
\midrule
TUSZ (4-Classes)    & Epilepsy & 2,891  & 4 & 19 & 6,000 & 1,734  & 578    & 579    \\ 
\midrule
PTB (2-Classes)     & Cardiopathy & 64,356 & 2 & 15 & 300  & 41,995 & 12,993 & 9,368  \\ 
\midrule
PTB-XL (4-Classes)  & Cardiopathy & 17,110 & 4 & 12 & 1,000 & 10,266 & 3,422  & 3,422  \\ 
\midrule
PTB-XL (5-Classes)  & Cardiopathy & 17,110 & 5 & 12 & 1,000 & 10,266 & 3,422  & 3,422 

% \end{tabular}
% \end{table}
            \\ % 这里要换行才能添加bottom rule
            \midrule \midrule
            %\midrule \midrule
            \end{tabular} % 表尾
\begin{tablenotes}
\item For readers' convenience, the public URLs corresponding to these datasets are provided as follows: \\
(1) APAVA: \url{https://osf.io/jbysn/} \\
(2) ADFTD: \url{https://openneuro.org/datasets/ds004504} \\
(3) TUSZ: \url{https://isip.piconepress.com/projects/nedc/html/tuh_eeg/} \\
(4) PTB: \url{https://physionet.org/content/ptbdb/1.0.0/} \\
(5) PTB-XL: \url{https://physionet.org/content/ptb-xl/1.0.3/}
\end{tablenotes}
        \end{threeparttable} % 三栏式，可以在下面加标注
        } % 盒子
%\vspace{-1mm} % 表格内容和标题的距离
 %引用
%\vspace{-2mm} %表格和其下排版的距离
\end{table*} % 双栏中的跨栏
%\end{table} % 单栏
%\end{wraptable} %文字围绕表格
%\input{./Figures/tex/data}

\subsection{Datasets}
\label{dat_app}

\subsubsection{\textbf{Datasets Details}}
(1) \textbf{APAVA (2-Classes)} \cite{escudero2006analysis} is a public EEG dataset for Alzheimer's disease (AD) classification. It contains two classes: "Healthy Person" and "Alzheimer's disease (AD)".  Since the dataset does not provide one-to-one text pairs, we adopt its dataset description and task description as associated clinical semantics:  
\textit{Dataset description}: The APAVA dataset comprises 16-channel EEG recordings for distinguishing patients with AD from healthy control subjects.
\textit{Task description}: Given the EEG signal, predict whether the sample belongs to a subject diagnosed with Alzheimer's disease (AD) or a healthy control (HC) subject.

(2) \textbf{ADFTD (3-Classes)} \cite{miltiadous2023dataset} is a public EEG dataset for Alzheimer's disease classification. It contains three classes: "Healthy Person (HC)" , "Frontotemporal Dementia (FTD)", and "Alzheimer's disease (AD)" .  Since the dataset does not provide one-to-one text pairs, we adopt its dataset description and task description as associated clinical semantics:  
\textit{Dataset description}: The ADFTD dataset comprises 16-channel EEG recordings for differentiating patients with AD, FTD, and healthy control subjects.  
\textit{Task description}: Given the EEG signal, predict whether the sample belongs to a subject diagnosed with Alzheimer's disease (AD), Frontotemporal Dementia (FTD), or a healthy control (HC) subject.

(3) \textbf{PTB (2-Classes)} \cite{physiobank2000physionet} is a public ECG dataset for heart disease classification. It contains two classes: "Healthy Person (HC)"  and "Myocardial infarction (MI)".  
Since the dataset does not provide one-to-one text pairs, we adopt its dataset description and task description as associated clinical semantics:  
\textit{Dataset description}: The PTB dataset comprises 15-lead ECG recordings for classifying and diagnosing cardiac conditions, including healthy controls and myocardial infarction patients.  
\textit{Task description}: Given the ECG signal, predict whether the sample belongs to a healthy control (HC) or a patient with myocardial infarction (MI).

(4) \textbf{PTB-XL} \cite{wagner2020ptb} is a large-scale public 12-lead ECG dataset for heart disease diagnosis.  
\textbf{PTB-XL (4-Classes)} consists of coarse-grained labels: "Abnormal ECG" , "Borderline ECG", "Normal ECG" , and "Otherwise normal ECG" .  
\textbf{PTB-XL (5-Classes)} provides fine-grained labels: "Conduction Disturbance" , "Hypertrophy" , "Myocardial Infarction" , "ST-T Changes", and "Normal ECG" .  
PTB-XL dataset provides clinical 12-lead ECGs and their corresponding reports. The clinical reports are automatically generated by the
machine and have no diagnosis revealed.  Since the dataset provides one-to-one clinical report, we adopt these clinical records as associated textual semantics. 
%An example is shown in Figure \ref{fig:data_app_2}.

(5) \textbf{TUSZ} \cite{shah2018temple} is a large-scale EEG dataset capturing brain electrical activity across 19 channels for epilepsy diagnosis.  
\textbf{TUSZ (2-Classes)} provides coarse-grained labels: "Normal EEG"  and "Abnormal EEG" .  
\textbf{TUSZ (4-Classes)} provides fine-grained labels, further categorizing abnormal EEG into four seizure types: "Combined focal (CF) seizures" , "Generalized non-specific (GN) seizures" , "Absence (AB) seizures" , and "Combined tonic (CT) seizures".  
TUSZ contains 19-channel EEG recordings along with patient-level clinical records for each diagnostic session, including clinical history, medications, and other relevant notes. Therefore, we adopt these records as the associated textual information for each EEG sample.
%An example is shown in Figure \ref{fig:data}.

\subsubsection{\textbf{Datasets Split}}
Table \ref{tab:data_app} provides the train-validation-test splits for all the datasets. The splits for the APAVA, ADFTD, and PTB datasets follow the previous work \cite{Medformer}, while the PTB-XL and TUSZ datasets employ a 60\%-20\%-20\% splitting strategy.

\subsubsection{\textbf{Data Pre-processing}}
For the data preprocessing of the APAVA, ADFTD, PTB, PTB-XL dataset and TUSZ datasets, we follow previous work \cite{Medformer}. 
\end{document}